\documentclass[format=acmsmall, review=false, screen=true]{acmart}

\usepackage{booktabs} 
\usepackage{multirow}
\usepackage{adjustbox}
\usepackage{rotating}
\usepackage{threeparttable}
\usepackage{subfigure}
\usepackage{xcolor,colortbl}
\usepackage{longtable}

\setcopyright{rightsretained}

\begin{document}

\title{Fake News Early Detection: An Interdisciplinary Study}

\author{Xinyi Zhou}
\email{zhouxinyi@data.syr.edu}
\author{Atishay Jain}
\email{atjain@syr.edu}
\author{Vir V. Phoha}
\email{vvphoha@syr.edu}
\author{Reza Zafarani}
\email{reza@data.syr.edu}

\affiliation{%
  \department{EECS Department}
  \institution{Syracuse University}
  \city{Syracuse}
  \state{NY}
  \postcode{13244}
  \country{USA}
  }

\begin{abstract}
Massive dissemination of fake news and its potential to erode democracy has increased the demand for accurate fake news detection. Recent advancements in this area have proposed novel techniques that aim to detect fake news by exploring how it propagates on social networks. Nevertheless, to detect fake news at an early stage, i.e., when it is published on a news outlet but not yet spread on social media, one cannot rely on news propagation information as it does not exist. Hence, there is a strong need to develop approaches that can detect fake news by focusing on news content. In this paper, a theory-driven model is proposed for fake news detection. The method investigates news content at various levels: lexicon-level, syntax-level, semantic-level and discourse-level. We represent news at each level, relying on well-established theories in social and forensic psychology. Fake news detection is then conducted within a supervised machine learning framework. As an interdisciplinary research, our work explores potential fake news patterns, enhances the interpretability in fake news feature engineering, and studies the relationships among fake news, deception/disinformation, and clickbaits. Experiments conducted on two real-world datasets indicate the proposed method can outperform the state-of-the-art and enable fake news early detection when there is limited content information. 
\end{abstract}




\maketitle

\section{Introduction}
\begin{figure}[t]
    \centering
    \includegraphics[scale=0.52]{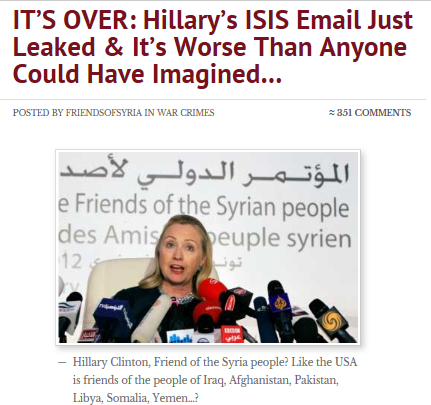} \qquad
    \includegraphics[scale=0.62]{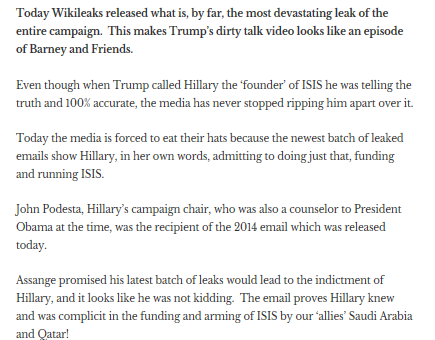}
    \caption{Fake News\protect\footnotemark. (1) This fake news story originally published on Ending the Fed has got $\sim$754,000 engagements in the final three months of the 2016 U.S. presidential campaign, which is the top-three-performing fake election news story on Facebook~\cite{silverman2016analysis}; (2) It is a fake news story with clickbait.}
    \label{fig:fakenews}
\end{figure}

Fake news is now viewed as one of the greatest threats to democracy and journalism~\cite{reza2019tutorial}. 
The reach of fake news was best highlighted during the critical months of the 2016 U.S. presidential election campaign, where the top twenty frequently-discussed fake election stories (see Figure \ref{fig:fakenews} for an example) generated 8,711,000 shares, reactions, and comments on Facebook, which is larger than the total of 7,367,000 for the top twenty most-discussed election stories posted by 19 major news websites~\cite{silverman2016analysis}. Our economies are not immune to the spread of fake news either, with fake news being connected to stock market fluctuations and massive trades. For example, fake news claiming that Barack Obama was injured in an explosion wiped out \$130 billion in stock value~\cite{rapoza2017can}. 

Meanwhile, humans have been proven to be not proficient in differentiating between truth and falsehood when overloaded with deceptive information. Studies in social psychology and communications have demonstrated that human ability to detect deception is only slightly better than chance: typical accuracy rates are in the range of 55\%-58\%, with a mean accuracy of 54\% over 1,000 participants in over 100 experiments~\cite{rubin2010deception}. Many expert-based (e.g., PolitiFact\footnote{\url{https://www.politifact.com/}} and Snope\footnote{\url{https://www.snopes.com/}}) 
and crowd-sourced (e.g., Fiskkit\footnote{\url{http://www.fiskkit.com/}} and TextThresher~\cite{zhang2018structured}) manual fact-checking websites, tools and platforms thus have emerged to serve the public on this matter\footnote{Comparison among common fact-checking websites is provided in \cite{zhou2018survey} and a comprehensive list of fact-checking websites is available at \url{https://reporterslab.org/fact-checking/}.}. Nevertheless, manual fact-checking does not scale well with the volume of newly created information, especially on social media~\cite{zafarani2014social}. Hence, automatic fake news detection has been developed in recent years, where current methods can be generally grouped into \textit{content-based} and \textit{propagation-based} methods.

\footnotetext{Direct source: \url{https://bit.ly/2uE5eaB}}

Content-based fake news detection aims to detect fake news by analyzing the content of news articles. Within a machine learning framework, researchers often detect fake news relying on either latent (via neural networks)~\cite{wang2018eann,zhou2019safe} or non-latent (usually hand-crafted) features~\cite{ciampaglia2015computational,shi2016discriminative,perez2017automatic,sitaula2019credibility}  of the content (see Section 2 for details). 
Nevertheless, in all such techniques, fundamental theories in social and forensic psychology have not played a significant role. Such theories can significantly improve fake news detection by highlighting some potential fake news patterns and facilitating explainable machine learning models for fake news detection~\cite{du2019techniques}. For example, \textit{Undeutsch hypothesis}~\cite{undeutsch1967beurteilung} states that a fake statement differs in writing style and quality from a true one. Such theories, as will be summarized in Section \ref{subsec:review_deception_clickbait}, can refer to either \textit{deception/disinformation}~\cite{undeutsch1967beurteilung,johnson1981reality,zuckerman1981verbal,mccornack2014information}, i.e., information that is intentionally and verifiably false, or \textit{clickbaits}~\cite{loewenstein1994psychology}, the headlines whose main purpose is to attract the attention of readers and encourage them to click on a link to a particular webpage~\cite{zhou2018survey}.
Compared to existing features, relying on such theories allows to introduce features that are explainable, can help the public well understand fake news, and help explore the relationships among fake news, deception/disinformation and clickbaits. Theoretically, deception/disinformation is a more general concept which includes fake news articles, fake statements, fake reviews, etc. Hence the characteristics attached to deception/disinformation might or might not be consistent with that of fake news, which motivates us to explore the relationships between fake news and types of deception. Meanwhile, clickbaits have been shown to be closely correlated to fake news~\cite{chen2015misleading}. The fake election news story in Figure \ref{fig:fakenews} is an example of a fake news story with a clickbait. When fake news meets clickbaits, we observe news articles that can attract eyeballs but are rarely news worthy~\cite{zhou2018survey}. Unfortunately, clickbaits help fake news attract more clicks (i.e., visibility) and further gain public trust, as indicated by the \textit{attentional bias}~\cite{macleod1986attentional}, which states that the public trust to a certain news article will increase with more exposure, as facilitated by clickbaits. On the other hand, while news articles with clickbaits are generally unreliable, not all such news articles are fake news, which motivates to explore the relationships between fake news and clickbait.

Unlike content-based fake news detection, propagation-based fake news detection aims to detect fake news by exploring how news propagates on social networks. Propagation-based methods have gained recent popularity where novel models have been proposed exhibiting reasonable performance~\cite{jin2016news,zhou2019network,castillo2011information,shu2019beyond,ruchansky2017csi,gupta2018cimtdetect,zhang2018fake}. 
However, propagation-based methods face a major challenge when detecting fake news. Within a life-cycle of any news article, there are three basic stages: being created, being published on news outlet(s), and being spread on social media (medium)~\cite{zhou2018survey}. Propagation-based models relying on social context information are difficult to be applied in predicting fake news before its third stage, i.e., before fake news has been propagated on social media. To detect fake news at an early stage, i.e., when it is published on a news outlet but has not yet spread on social media sites, in order to take early actions for fake news intervention (i.e., \textit{fake new early detection}) motivates us to deeply mine news content.
Such early detection is particularly crucial for fake news as more individuals become exposed to some fake news, the more likely they may trust it~\cite{boehm1994validity}. Meanwhile, it has been demonstrated that it is difficult to correct one's cognition after fake news has gained their trust (i.e., \textit{Semmelweis reflex}~\cite{balint2009semmelweis},  \textit{confirmation bias}~\cite{nickerson1998confirmation}, and \textit{anchoring bias}~\cite{tversky1974judgment}). 


In summary, current development in fake news detection strongly motivates the need for techniques that deeply mine news content and rely less on how fake news propagates. Such techniques should investigate how social and forensic theories can help detect fake news for interpretablity reasons. Here, we aim to address these challenges by developing a theory-driven fake news detection model that concentrates on news content to be able to detect fake news before it has been propagated on social media. The model represents news articles by a set of manual features, which capture both content structure and style across language levels (i.e., lexicon-level, syntax-level, semantic-level and discourse-level) via conducting an interdisciplinary study. Features are then utilized for fake news detection within a supervised machine learning framework. The specific contributions of this paper are as follows:

\vspace{-1mm}
\begin{enumerate}
    \item We propose a model that enables fake news early detection. By solely relying on news content, the model allows to conduct detection when fake news has been published on a news outlet while has not been disseminated on social media. Experimental results on real-world datasets validate the model effectiveness when limited news content information is available.
    \item We conduct an interdisciplinary fake news study by broadly investigating social and psychological theories, which presents a systematic framework for fake news feature engineering. Within such framework, each news article is represented respectively at the lexicon-, syntax-, semantic- and discourse-level within language. Compared to latent features, such features can enhance model interpretability, help fake news pattern discovery, and help the public better understand fake news. 
    \item We explore the relationships among fake news, types of deception, and clickbaits. By empirically studying their characteristics in, e.g., content quality, sentiment, quantity and readability, some news patterns unique to fake news or shared with deception or clickbaits are revealed.
\end{enumerate}

\vspace{-1mm}
The rest of this paper is organized as follows. Literature review is presented in Section \ref{sec:review}. The proposed model is specified in Section \ref{sec:method}. In Section \ref{sec:experiment}, we evaluate the performance of our model on two real-world datasets. Section \ref{sec:conclusion} concludes the paper.

\section{Related Work}
\label{sec:review}

Our work is mainly related to the detection of fake news, with the investigation of its characteristics as well as that of types of deception/disinformation and clickbaits. Next, we review the development of fake news detection in Section \ref{subsec:review_fakeNews}, and summarize the characteristics of fake news, deception, and clickbait in Section \ref{subsec:review_deception_clickbait}.

\subsection{Fake News Detection}
\label{subsec:review_fakeNews}

Depending on whether the approaches detect fake news by exploring its content or by exploring how it propagates on social networks, current fake news detection studies can be generally grouped into content-based and propagation-based methods. We review recent advancements on both fronts.

\subsubsection{Content-based Fake News Detection} 
\label{subsubsec:review_fnContent}

In general, current content-based approaches detect fake news by representing news content within different frameworks. Such representation of news content can be from the perspective of (I) knowledge or (II) style, or can be a (III) latent representation.\vspace{2mm}

\noindent \textit{I. Knowledge} is often defined as a set of SPO (Subject, Predicate, Object) tuples extracted from text. An example of such knowledge (i.e., SPO tuples) is (DonaldTrump, Profession, President) for the sentence ``Donald Trump is the president of the U.S.''
Knowledge-based fake news detection usually develops link prediction algorithms~\cite{ciampaglia2015computational,shi2016discriminative} with the goal of directly evaluating news authenticity by comparing (inferring) the knowledge extracted from to-be-verified news content with that within a Knowledge Graph (KG) such as Knowledge Vault~\cite{dong2014knowledge}. KGs, often regarded as ground truth datasets, contain massive manually-processed relational knowledge from the open Web. However, one has to face various challenges within such a framework. Firstly, KGs are often far from \textit{complete}, often demanding further post-processing approaches for knowledge inference~\cite{nickel2016review}. Second, news, as newly received or noteworthy information especially about recent events, demands knowledge to be \textit{timely} within KGs. Third, knowledge-based approaches can only evaluate if the to-be-verified news article is false instead of being \textit{intentionally} false, where the former refers to false news while the latter refers to fake news~\cite{zhou2018survey}.\vspace{2mm}

\noindent \textit{II. Style} is a set of self-defined [non-latent] machine learning features that can represent fake news and differentiate it from the truth~\cite{zhou2018survey}.
For example, such style features can be word-level statistics based on TF-IDF, $n$-grams and/or LIWC features~\cite{perez2017automatic,potthast2017stylometric,castelo2019topic}, and rewrite-rule statistics based on TF-IDF~\cite{perez2017automatic}. 
Though these style features can be comprehensive in detecting fake news, their selection or extraction is driven by experience that are rarely supported by fundamental theories across disciplines. An example of such a machine learning framework is the interesting study by Rubin and Lukoianova, which identifies fake news by combining rhetorical structures with vector space model~\cite{rubin2015truth}.\vspace{2mm}

\noindent \textit{III. Latent features} represent news articles via automatically generated features often obtained by matrix/tensor factorization or deep learning techniques, e.g., Text-CNN~\cite{wang2018eann,khan2019benchmark,zhou2019safe}.
Though these latent features can perform well in detecting fake news, they are often difficult to be comprehended, which brings challenges to promote the public's understanding of fake news. 

\subsubsection{Propagation-based Fake News Detection}
\label{subsubsec:review_fnPropagation}

Propagation-based fake news detection further utilizes social context information to detect fake news, e.g., how fake news propagates on social networks, who spreads the fake news, and how spreaders connect with each other~\cite{monti2019fake}. 

A direct way of presenting news propagation is using a \textit{news cascade} - a tree structure presenting post-repost relationships for each news article on social media, e.g., tweets and retweets on Twitter~\cite{zhou2018survey,castillo2011information}. Based on news cascades, for example, Wu et al.~\cite{wu2015false} extend news cascades by introducing user roles (i.e., opinion leaders or normal users), stance (i.e., approval or doubt) and sentiments expressed in user posts. By assuming that the overall structure of fake news cascades differs from true ones, the authors develop a random walk graph kernel to measure the similarity among news cascades and detect fake news based on such similarity. Liu and Wu model news cascades as multivariate time series. Based on that, fake news is detected by incorporating both Recurrent Neural Network (RNN) and Convolutional Neural Network (CNN)~\cite{liu2018early}.

In addition to news cascades, some self-defined graphs that can indirectly represent news propagation on social networks are also constructed for fake news detection. Jin et al.~\cite{jin2016news} build a stance graph based on user posts, and detect fake news by mining the stance correlations within a graph optimization framework. By exploring relationships among news articles, publishers, users (spreaders) and user posts, PageRank-like algorithm~\cite{gupta2012evaluating}, matrix and tensor factorization~\cite{gupta2018cimtdetect,shu2019beyond}, or RNN~\cite{ruchansky2017csi,zhang2018fake} have been developed for fake news detection. 

While remarkable progress has been made, to detect fake news at an early stage, i.e.,  when it is published on a news outlet and before it has been spread on any social media, one cannot rely on social context information and in turn, propagation-based methods, as only limited or no social context information is available at the time of posting for fake news articles. Hence, to design a fake news early detection technique, we solely rely on mining news content. 

\subsection{Fake News, Deception and Clickbait Characteristics}
\label{subsec:review_deception_clickbait}

We review the studies that reveal the characteristics of fake news, deception and clickbait respectively in Section \ref{subsubsec:review_fakeNews} to Section \ref{subsubsec:review_clickbait}.

\subsubsection{Fake News}
\label{subsubsec:review_fakeNews}

Most current studies focus on investigating the patterns and characteristics in the propagation of fake news compared to that of the truth~\cite{parikh2019towards}. 
Vosoughi et al. investigate the differential diffusion of true and fake news stories distributed on Twitter from 2006 to 2017, where the data comprise $\sim$126,000 stories tweeted by $\sim$3 million people more than 4.5 million times~\cite{vosoughi2018spread}. The authors discover that fake news diffuses significantly farther, faster, more broadly, and can involve more individuals than the truth. They observe that these effects are more pronounced for fake political news than for fake news about terrorism, natural disasters, science, urban legends, or financial information. Recently, Zhou and Zafarani reveal that fake news spreaders often form a denser social network compared to true news spreaders~\cite{zhou2019network}. 

\subsubsection{Deception}
\label{subsubsec:review_deception}

Deception (disinformation) is the information that is intentionally false~\cite{zhou2018survey}. Deception has various forms, where fake (deceptive) statements and justifications are referred by most studies. Fundamental theories in psychology and social science have revealed some linguistic cues when a person lies compared to when he or she tells the truth. For example, \textit{Undeutsch hypothesis}~\cite{undeutsch1967beurteilung} states that a statement based on a factual experience differs in content style and quality from that of fantasy; \textit{reality monitoring}~\cite{johnson1981reality} indicates that actual events are characterized by higher levels of sensory-perceptual information; 
\textit{four-factor theory}~\cite{zuckerman1981verbal} reveals that lies are expressed differently in terms of emotions and cognitive processes from truth; and
\textit{information manipulation theory}~\cite{mccornack2014information} validates that extreme information quantity often exists in deception. 

\subsubsection{Clickbait}
\label{subsubsec:review_clickbait}

Clickbait is the headlines whose main purpose is to attract the attention of readers and encourage them to click on a link to a particular Web page. Examples of clickbait are ``33 Heartbreaking Photos Taken Just Before Death'', ``You Won't Believe Obama Did That No President Has Ever Done!'' and ``IT'S OVER: Hillary's ISIS Email Just Leaked \& It's Worse Than Anyone Could Have Imagined...'' (Figure \ref{fig:fakenews}). 
To achieve the purpose, clickbait creators make great efforts to produce an \textit{information gap}~\cite{loewenstein1994psychology} between the headlines and individuals' knowledge. Such information gaps produce the feeling of deprivation labeled curiosity, which motivates individuals to obtain the missing information to reduce such feeling.

\section{Methodology}
\label{sec:method}

In this section, we detail the proposed method of predicting fake news. Before further elaboration, we formally define the target problem as below:

\vspace{0.5em}
\noindent \textbf{Problem Definition.}
Assume a to-be-verified news article can be represented as a feature vector $\mathbf{f} \in \mathbb{R}^n$, where each entry of $\mathbf{f}$ is a linguistic machine learning feature. The task to classify the news article based on its content representation is to identify a function $\mathcal{A}$, such that 
$ \mathcal{A}: \mathbf{f} \xrightarrow{TD} \hat{y} $,
where $\hat{y} \in \{0,1\}$ is the predicted news label; 1 indicates that the news article is predicted as fake news and 0 indicates it is true news. $TD= \{ (\mathbf{f}^{(k)},y^{(k)}): \mathbf{f}^{(k)} \in \mathbb{R}^n, y^{(k)} \in \{0,1\}, k \in \mathbb{N}_+ \}$ is the training dataset. The training dataset helps estimate the parameters within $\mathcal{A}$ and consists of a set of news articles represented by the same set of features ($\mathbf{f}^{(k)}$) with known news labels ($y^{(k)}$).

\vspace{0.5em}
Within the aforementioned traditional supervised learning framework, an explainable and well-performed  method of predicting fake news relies on (1) the way that a news article is represented ($\mathbf{f}$), and (2) the classifier used to predict fake news ($\mathcal{A}$). Next, we will specify each in Section \ref{subsec:method_representation} and Section \ref{subsec:method_classification}, respectively.

\subsection{News Representation}
\label{subsec:method_representation}

As suggested by \textit{Undeutsch hypothesis}~\cite{undeutsch1967beurteilung}, fake news potentially differs in \textit{writing style} from true news. Thus, we represent news content by capturing its writing style respectively at lexicon-level (Sec. \ref{subsubsec:lexicon}), syntax-level (Sec. \ref{subsubsec:syntax}), semantic-level (Sec. \ref{subsubsec:semantic}) and discourse-level (Sec. \ref{subsubsec:discourse}). 

\subsubsection{Lexicon-level} 
\label{subsubsec:lexicon}
To capture news writing style at lexicon-level, we investigate the frequency of words being used in news content, where such frequency can be simply obtained by a \textit{Bag-Of-Word (BOW)} model. However, BOW representation can only capture the absolute frequencies of terms within a news article rather than their \textit{relative (standardized) frequencies} which have accounted for the impact of content length (i.e., the overall number of words within the news content); the latter is more representative when extracting writing style features based on the words or topics that authors prefer to use or involve. 
Therefore, we first use a standardized BOW model to represent the writing style of each news article at the lexicon-level. 
Mathematically, assume a corpus contains $p$ news articles $M=\{m_1,m_2,\cdots,m_p\}$ with a total of $q$ words $W=\{w_1,w_2,\cdots,w_q\}$. $x^i_j$ denotes the number of $w_j$ appearing in $m_i$. Then the standardized frequency of $w_j$ for news $m_i$ is ${x^i_j}/{\sum_{j=1}^{q}x^i_j}$.

\subsubsection{Syntax-level} 
\label{subsubsec:syntax}
\begin{figure}[t]
    \centering
    \includegraphics[width=0.85\textwidth]{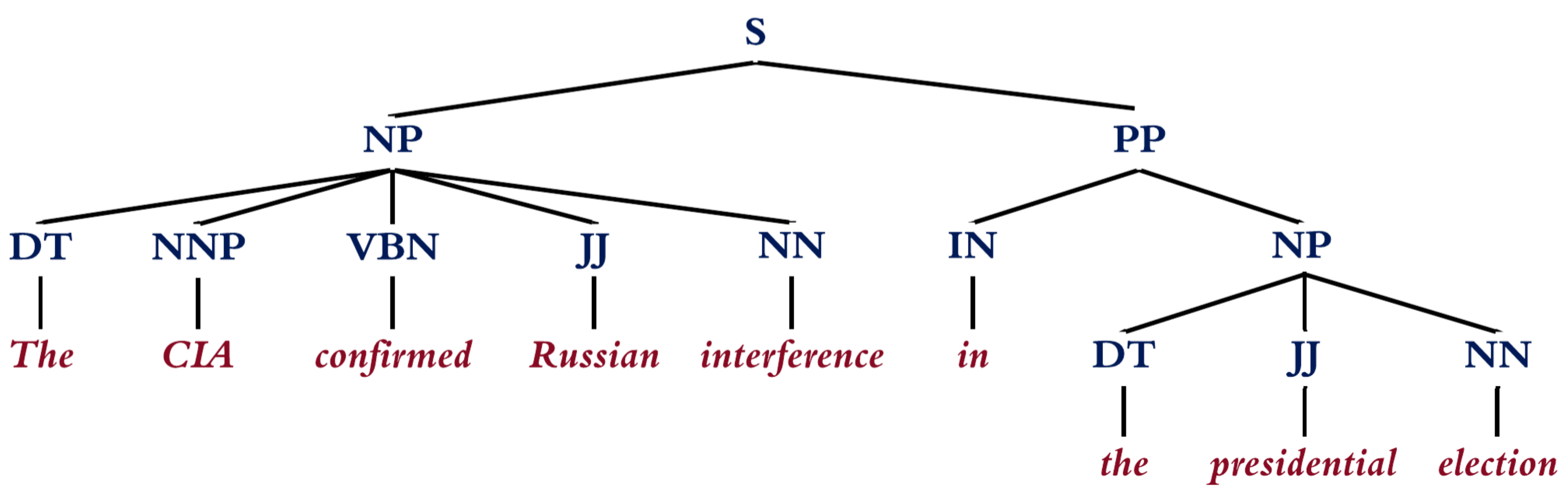}
    \caption{PCFG Parsing Tree for the sentence ``The CIA confirmed Russian interference in the presidential election'' within a fake news article. The rewrite rules of this sentence should be the following: S $\rightarrow$ NP PP, NP $\rightarrow$ DT NNP VBN JJ NN, PP $\rightarrow$ IN NP, NP $\rightarrow$ DT JJ NN, DT $\rightarrow$ `the', NNP $\rightarrow$ `CIA', VBN $\rightarrow$ `confirmed', JJ $\rightarrow$ `Russian', NN $\rightarrow$ `inference', IN $\rightarrow$ `in', JJ $\rightarrow$ `presidential' and NN $\rightarrow$ `election'. }
    \label{fig:cfg}
\end{figure}

Syntax-level style features can be further grouped into shallow syntactic features and deep syntactic features~\cite{feng2012syntactic}, where the former investigates the frequency of \textit{Part-Of-Speech (POS) tags} (e.g., nouns, verbs and determiners) and the latter investigates the frequency of productions (i.e., \textit{rewrite rules}). The rewrite rules of a sentence within a news article can be obtained based on Probability Context Free Grammar (PCFG) parsing trees. An illustration is shown in Figure \ref{fig:cfg}. Here, we also compute the frequencies of POS tags and rewrite rules of a news articles in a relative (standardized) way, which removes the impact of news content length (i.e., instead of denoting the $i$-th word, $w_i$ defined in last section here indicates the $i$-th POS tag or rewrite rule).

\subsubsection{Semantic-level}
\label{subsubsec:semantic}
Style features at semantic-level investigate some psycho-linguistic attributes, e.g., sentiments, expressed in news content. Such attributes defined and assessed in our work are basically inspired by well-established fundamental theories initially developed in forensic- and social-psychology. 
As specified in Section \ref{subsec:review_deception_clickbait}, most these theories are not accurately developed fake news, but for deception/disiformation or clickbait that includes or closely relates to fake news.
We detail our feature engineering at semantic-level by separating news content as headlines and body-text, where clickbait-related attributes target news headlines and disinformation-related ones are mainly concerned with news body-text. A detailed list of semantic-level features defined and selected in our study is provided in Appendix \ref{app:semantic_features}.

\vspace{0.5em}
\noindent \textbf{ClickBait-related Attributes (CBAs).}
Clickbaits have been suggested to have a close relationship with fake news, where clickbaits help enhance click-through rates for fake news articles and in turn, further gain public trust~\cite{macleod1986attentional}. 
Hence, we aim to extract a set of features that can well represent clickbaits to capture fake news headlines.  
We evaluate news headlines from the following four perspectives.

\paragraph{A. General Clickbait Patterns}
We have utilized two public  dictionaries\footnote{\url{https://github.com/snipe/downworthy}} that provide some common clickbait phrases and expressions such as ``can change your life'' and ``will blow your mind''~\cite{gianotto2014downworthy}. A general way of representing news headlines based on these dictionaries is to verify if a news headline contains any of the common clickbait phrases and/or expressions listed, or how frequent such common clickbait phrases and/or expressions are in the news headline. Due to the length of news headlines, here the frequency of each clickbait phrase or expression is not considered in our feature set as it leads to many zeros in our feature matrix. Such dictionaries have been successfully applied in clickbait detection~\cite{potthast2016clickbait,chakraborty2017tabloids,jaidka2018predicting}. 

\paragraph{B. Readability} Psychological research has indicated that a clickbait attracts public eyeballs and encourages clicking behavior by creating an \textit{information gap} between the knowledge within the news headline and individuals' existing knowledge~\cite{loewenstein1994psychology}. Such information gap has to be produced on the basis that the readers have understood what the news headline expresses. Therefore, we investigate the readability of news headlines by employing several well-established metrics developed in education, e.g., Flesch Reading Ease Index (FREI), Flesch-Kincaid Grade Level (FKGL), Automated Readability Index (ARI), Gunning Fog Index (GFI), and Coleman-Liau Index (CLI). We also separately consider and include as features the parameters within these metrics, i.e., the number of characters, syllables, words, and long (complex) words.

\paragraph{C. Sensationalism} To produce an information gap~\cite{loewenstein1994psychology}, further attract public attention, and encourage users to click, expressions with exaggeration and sensationalism are common in clickbaits. As having been suggested in clickbait dictionaries~\cite{gianotto2014downworthy}, clickbait creators prefer to use ``can change your life'' which might actually ``not change your life in any meaningful way''; or use ``will blow your mind'' to replace ``might perhaps mildly entertain you for a moment'', where the former rarely happens compared to the latter and thus produces the information gap. We evaluate the sensationalism degree of a news headline from the following aspects.
        \begin{itemize}
        \item \textit{Sentiment.} Extreme sentiment expressed in a news headline is assumed to indicate a higher degree of sensationalism. Hence, we measure the frequencies of positive words and negative words within a news headline by using LIWC, as well as the news headline sentiment polarity by computing the average sentiment scores of the words it contains.
        \item \textit{Punctuation}: Some punctuations can help express sensationalism or extreme sentiments, e.g., quotes (`...'), question (`?') and exclamation marks (`!'). Hence the frequencies of these three are also counted when representing news headlines.
        \item \textit{Similarity.} Similarity between the headline of a news article and its body-text is assumed to be positively correlated to the degree of \textit{relative} sensationalism expressed in the news headline~\cite{dong2019similarity}. Capturing such similarity requires firstly embedding the headline and body-text for each news article into the same space. To achieve this goal, we respectively utilize \textsc{word2vec}~\cite{mikolov2013efficient} model at the word-level and train \textsc{Sentence2Vec}~\cite{arora2016simple} model at the sentence-level, considering that one headline often refers to one sentence. For the headline or body-text containing more than one words or sentences, we compute the average of its word embedding (i.e., vectors) or sentence embedding. The similarity between a news headline and its body-text then can be computed based on various similarity measures, where we use cosine distance in experiments. To our best knowledge, similarity between the headlines and their body-text is first captured in such way.
        \end{itemize}

\paragraph{D. News-worthiness} While click-baits can attract eyeballs they are rarely newsworthy with (I) low quality and (II) high informality~\cite{zhou2018survey}. We capture both characteristics in news:
        \begin{itemize}
        \item \textit{I. Quality}: The title of high quality news articles is often a \textit{summary} of the whole news event described in the body-text~\cite{dong2019similarity}. To capture this property, one can assess the similarity between the headline of a news article and its body-text, which has been already captured when analyzing sensationalism. Secondly, such titles should be a \textit{simplified} summary of the whole news event described in body-text, where meaningful words should occupy its main proportion~\cite{chakraborty2016stop}. From this perspective, the frequencies of content words, function words, and stop words within each news headline are counted and included as features. 
        \item \textit{II. Informality}: LIWC~\cite{pennebaker2015development} provides five dimensions to evaluate such informality of language: 
        (1) \textit{swear words} (e.g., `damn'); 
        (2) \textit{netspeaks} (e.g., `btw' and `lol');  
        (3) \textit{assents} (e.g., `OK'); 
        (4) \textit{nonfluencies} (e.g., `er', `hm', and `umm'); and 
        (5) \textit{fillers} (e.g., `I mean' and `you know'). 
        Hence, we measure the informality for each news headline by investigating its word or phrase frequencies within every dimension and include them as features.
        \end{itemize}

\vspace{0.5em}
\noindent \textbf{Disinformation-related Attributes (DIAs).}
Deception/disinformation is a more general concept compared to fake news, which additionally includes fake statements, fake reviews, and the like~\cite{brunton2013spam}. 
Here we aim to extract a set of features inspired by disinformation-related theories such as \textit{Undeutsch hypothesis}~\cite{undeutsch1967beurteilung}, \textit{reality monitoring}~\cite{johnson1981reality}, \textit{four-factor theory}~\cite{zuckerman1981verbal}, and
\textit{information manipulation theory}~\cite{mccornack2014information} to represent news content. Such features are with respect to:

\paragraph{A. Quality}: In addition to writing style, \textit{Undeutsch hypothesis}~\cite{undeutsch1967beurteilung} states that a fake statement also differs in quality from a true one. Here, we evaluate news quality from three perspectives:

\begin{itemize}
    \item \textit{Informality}: Basically, the quality of a news article should be negatively correlated to its informality. As having been specified, LIWC~\cite{pennebaker2015development} provides five dimensions to evaluate the informality of language. Here, we investigate the word or phrase numbers (proportions) on each dimension within news content (as apposed to headline) and include them as features.
    
    \item \textit{Diversity}: Such quality can possibly be assessed by investigating the number (proportion) of unique (non-repeated) words, content words, nouns, verbs, adjectives and adverbs being used in news content. We compute and include them as features as well.
    
    \item \textit{Subjectivity}: When a news article becomes hyperpartisan and biased, its quality should also be considered to be lower compared with those that maintain objectivity~\cite{potthast2017stylometric}. Benefiting from the work done by Recasens et al.~\cite{recasens2013linguistic}, which provides the corpus of biased lexicons, here we evaluate the subjectivity of news articles by counting their number (proportion) of biased words. On the other hand, factive verbs (e.g., `observe')~\cite{hooper1975assertive} and report verbs (e.g., `announce')~\cite{recasens2013linguistic}, as the opposite of biased ones, their numbers (proportions) are also included in our feature set, which are negatively correlated to content subjectivity.
\end{itemize}

\paragraph{B. Sentiment} Sentiment expressed within news content is suggested to be different within fake news and true news~\cite{zuckerman1981verbal}. Here, we evaluate such sentiments for each news article by measuring the number (proportion) of positive words and negative words, as well as its sentiment polarity.

\paragraph{C. Quantity} \textit{Information manipulation theory}~\cite{mccornack2014information} reveals that extreme information quantity (too much or too little) often exists in deceptive content. We assess such quantity for each news article at character-level, word-level, sentence-level and paragraph-level, respectively, i.e., the overall number of characters, words, sentences and paragraphs; and the average number of characters per word, words per sentence, sentences per paragraph.

\paragraph{D. Specificity} Fictitious stories often differs in cognitive and perceptual processes, as indicated by \textit{reality monitoring}~\cite{johnson1981reality} and \textit{four-factor theory}~\cite{zuckerman1981verbal}. Based on LIWC dictionary~\cite{pennebaker2015development}, for \textit{cognitive processes}, we investigate the frequencies of terms related to (1) \textit{insight} (e.g., `think'), (2) \textit{causation} (e.g., `because'), (3) \textit{discrepancy} (e.g., `should'), (4) \textit{tentative language} (e.g., `perhaps'), (5) \textit{certainty} (e.g., `always') and (6) \textit{differentiation} (e.g., `but' and `else'); for \textit{perceptual processes}, we investigate the frequencies of terms referring to vision, hearing, and feeling.

\begin{figure}[t]
    \centering
    \includegraphics[width=0.9\textwidth]{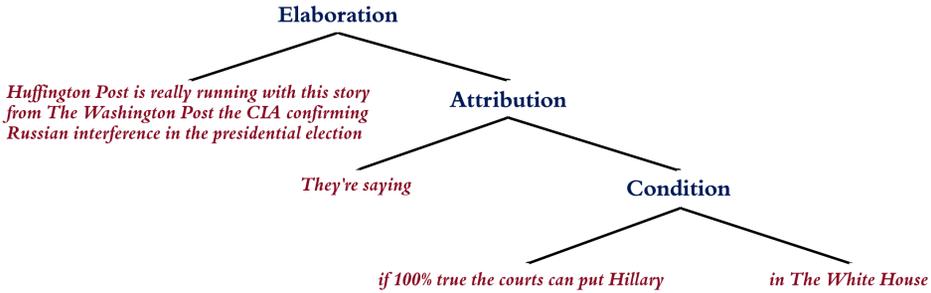}
    \caption{Rhetorical Structure for the partial content ``Huffington Post is really running with this story from The Washington Post about the CIA confirming Russian interference in the presidential election. They're saying if 100\% true, the courts can PUT HILLARY IN THE WHITE HOUSE!'' within a fake news article. Here, one elaboration, attribution and condition rhetorical relationships exist.}
    \label{fig:rst}
\end{figure}

\subsubsection{Discourse-level}
\label{subsubsec:discourse}
We first extract style features at discourse-level by investigating the standardized frequencies of rhetorical relationships among phrases or sentences within a news article. Instead of denoting the $i$-th word, $w_i$ defined in Section \ref{subsubsec:lexicon} here indicates the $i$-th rhetorical relationship. Such relationships can be obtained through an RST parser\footnote{\url{https://github.com/jiyfeng/DPLP}}~\cite{ji2014representation}, where an example is given in Fig. \ref{fig:rst}. Specifically, for a given piece of content, its rhetorical relationships among phrases or sentence can form a tree structure, where each leaf node is a phrase or sentence while non-leaf node is the corresponding rhetorical relationship between two phrases or sentences.

\subsection{News Classification}
\label{subsec:method_classification}

We have detailed how each news article can be represented across language levels with computational features inspired by fundamental theories. These features then can be utilized by supervised classifiers that are widely-accepted and well-established, e.g., Logistic Regression (LR), Na\"ive Bayes (NB), Support Vector Machine (SVM), Random Forests (RF), and XGBoost~\cite{chen2016xgboost}, for fake news prediction.  
As classifiers perform best for machine learning settings they were initially designed for (i.e., \textit{no free lunch theorem}), it is illogical to determine algorithms that perform best for fake news detection~\cite{zhou2018survey}. Following the common machine learning setting, we experiment with each of the aforementioned classifier based on our features and experimental settings; results will be comprehensively presented by multiple classifiers performing the best~\cite{zhou2019network}.

\section{Experiments}
\label{sec:experiment}

We conduct empirical studies to evaluate the proposed model, where experimental setup is detailed in Section \ref{subsec:setup}, and the performance is presented and evaluated in Section \ref{subsec:performance}.

\subsection{Experimental Setup}
\label{subsec:setup}

Real-world datasets used in our experiments are specified in Section \ref{subsubsec:dataset} followed by baselines that our model will be compared to in Section \ref{subsubsec:baselines}.

\subsubsection{Datasets}
\label{subsubsec:dataset}
Our experiments are conducted on two well-established public benchmark datasets of fake news detection\footnote{ \url{https://github.com/KaiDMML/FakeNewsNet/tree/old-version}}~\cite{shu2019beyond,shu2018fakenewsnet,shu2019role}.
News articles in these datasets are collected from PolitiFact and BuzzFeed, respectively. 
Ground truth labels (\textit{fake} or \textit{true}) of news articles in both datasets are provided by fact-checking experts, which guarantees the quality of news labels (\textit{fake} or \textit{true}).
In addition to news content and labels, both datasets also provide massive information on social network of users involved in spreading true/fake news on Twitter containing (1) users and their following/follower relationships (\textit{user-user relationships}) and (2) how the news has been propagated (tweeted/re-tweeted) by Twitter users, i.e., \textit{news-user relationships}. 
Such information is valuable for our comparative studies.
Statistics on the two datasets are provided in Table~\ref{tab::datasets}.
Note that the original datasets are balanced with 50\% true news and 50\% fake news. As few reference studies have provided the actual ratio between true news and fake news, we design an experiment in Section \ref{subsubsec:newsDistribution} to evaluate our work within unbalanced datasets by controlling this ratio.

\begin{table}[t]
\small
\caption{Data Statistics}
\label{tab::datasets}
\begin{tabular}{lrr}
\toprule[1pt]
\multicolumn{1}{l}{\textbf{Data}} & \multicolumn{1}{c}{\textbf{PolitiFact}} & \multicolumn{1}{c}{\textbf{BuzzFeed}} \\ \hline
 \# Users & 23,865 & 15,257 \\ 
 \# News--Users & 32,791 & 22,779 \\ 
 \# Users--Users & 574,744 & 634,750 \\
 \# News Stories & 240 & 180 \\ 
 \# True News & 120 & 90 \\ 
 \# Fake News & 120 & 90 \\
  \bottomrule[1pt] 
\end{tabular}
\end{table}


\subsubsection{Baselines}
\label{subsubsec:baselines}

We compare the performance of the proposed method with several state-of-the-art fake news detection methods on the same datasets. These methods detect fake news by (1) analyzing news content (i.e., content-based fake news detection)~\cite{perez2017automatic}, or (2) exploring news dissemination on social networks (i.e., propagation-based fake news detection)~\cite{castillo2011information}, or (3) utilizing both information within news content and news propagation information~\cite{shu2019beyond}. 

\vspace{0.5em}
\noindent \textbf{I. P\'erez-Rosas et al.~\cite{perez2017automatic}} propose a comprehensive linguistic model for fake news detection, involving the following features:  (i) $n$-grams (i.e., uni-grams and bi-grams) and (ii) CFGs based on TF-IDF encoding; (iii) word and phrase proportions referring to all categories provided by LIWC; and (iv) readability. Features are computed and used to predict fake news within a supervised machine learning framework.

\vspace{0.5em}
\noindent \textbf{II. Castillo et al.~\cite{castillo2011information}} design features to exploit information from user profiles, tweets and propagation trees to evaluate news credibility within a supervised learning framework. Specifically, these features are based on (i) quantity, sentiment, hash-tag and URL information from user tweets, (ii) user profiles such as registration age, (iii) news topics through mining tweets of users, and (iv) propagation trees (e.g., the number of propagation trees for each news topic).


\vspace{0.5em}
\noindent \textbf{III. Shu et al.~\cite{shu2019beyond}} detect fake news by exploring and embedding the relationships among news articles, publishers and spreaders on social media. Specifically, such embedding involves (i) news content by using non-negative matrix factorization, (ii) users on social media, (iii) news-user relationships (i.e., user engagements in spreading news articles), and (iv) news-publisher relationships (i.e., publisher engagements in publishing news articles). Fake news detection is then conducted within a semi-supervised machine learning framework.

\vspace{0.5em} 
Additionally, fake news detection based on latent representation of news articles is also investigated in other studies. Compared to style features, such latent features are less explainable but have been empirically shown to be remarkably useful~\cite{oshikawa2018survey,wang2017liar}. Here we consider as baselines classifiers that use as features \textbf{(IV) \textsc{word2vec}}~\cite{mikolov2013efficient} and  \textbf{(V) \textsc{Doc2Vec}}~\cite{le2014distributed} embeddings of news articles.

\subsection{Performance Evaluation}
\label{subsec:performance}

In our experiments, each dataset is randomly divided into training and testing datasets with the ratio $0.8:0.2$. Several supervised classifiers have been used with five-fold cross-validation, among which SVM (with linear kernel), Random Forest (RF) and XGBoost\footnote{\url{https://github.com/dmlc/XGBoost}}~\cite{chen2016xgboost} perform best compared to the others (e.g., LR, Logistic Regression and NB, Na\"ive Bayes) within both our model and baselines.
The performance of the experiments are provided in terms of accuracy, precision, recall and $F_1$ scores.
In this section, we first present and evaluate the general performance of the proposed model by comparing it with baselines in Section \ref{subsubsec:generalPerform}.
As news content is represented at the lexicon, syntax, semantic and discourse levels, we evaluate the performance of the model within and across different levels in Section \ref{subsubsec:acrossLevel}. The detailed analysis at the semantic-level follows, which provides opportunities to investigate the potential and understandable patterns of fake news, as well as its relationships with deception/disinformation (Section \ref{subsubsec:deception}) and clickbaits (Section \ref{subsubsec:clickbait}).
Next, we assess the impact of news distribution on the proposed model in Section \ref{subsubsec:newsDistribution}.
Finally, we investigate the performance of the proposed method for fake news early detection in Section \ref{subsubsec:early}.

\subsubsection{General Performance in Predicting Fake News} \label{subsubsec:generalPerform}
Here, we provide the general performance of the proposed model in predicting fake and compare it with baselines. 
Results are presented in Table \ref{tab:generalPerformance}, which indicate that among baselines, (1) the propagation-based fake news detection model (\cite{castillo2011information}) can perform comparatively well compared to content-based ones (\cite{perez2017automatic,mikolov2013efficient,le2014distributed}); and (2) the hybrid model (\cite{shu2019beyond}) can outperform fake news detection models that use either news content or propagation information. 
Compared to the baselines, (3) our model [slightly] outperforms the hybrid model in predicting fake news, while not relying on propagation information. 
For fairness of comparison, we report the best performance of the methods that rely on supervised classifiers by using SVM, RF, XGBoost, LR and NB:  \textsc{word2vec} features~\cite{mikolov2013efficient} and features in the work of Perez-Rosas et al.~\cite{perez2017automatic} perform best with linear SVM; while features based on \textsc{Doc2Vec}~\cite{le2014distributed}, in the work of Castillo et al.~\cite{castillo2011information}, and in our work perform best by using XGBoost.

\begin{table}[t]
\small
\caption{General Performance of Fake News Detection Models\protect\footnotemark[10].
Among the baselines, (1) the propagation-based model (\cite{castillo2011information}) can perform relatively well compared to content-based ones (\cite{perez2017automatic,mikolov2013efficient,le2014distributed}); and (2) the hybrid model (\cite{shu2019beyond}) can outperform both types of techniques. 
Compared to the baselines, (3) our model [slightly] outperforms the hybrid model and can outperform the others in predicting fake news.}
\label{tab:generalPerformance}
\begin{tabular}{rcccc|cccc}
\toprule[1pt]
\multirow{2}{*}{\textbf{Method}} & \multicolumn{4}{c|}{\textbf{PolitiFact}} & \multicolumn{4}{c}{\textbf{BuzzFeed}} \\ \cline{2-9}
 & \textbf{Acc.} & \textbf{Pre.} & \textbf{Rec.} & $\mathbf{F_1}$ & \textbf{Acc.} & \textbf{Pre.} & \textbf{Rec.} & $\mathbf{F_1}$ \\ \hline
\textbf{Perez-Rosas et al.~\cite{perez2017automatic}} & .811 & .808 & .814 & .811 & .755 & .745 & .769 & .757 \\
{$n$-grams+TF-IDF} & .755 & .756 & .754 & .755 & .721 & .711 & .735 & .723 \\
{CFG+TF-IDF} & .749 & .753 & .743 & .748 & .735 & .738 & .732 & .735 \\
{LIWC} & .645 & .649 & .645 & .647 & .655 & .655 & .663 & .659 \\
{Readability} & .605 & .609 & .601 & .605 & .643 & .651 & .635 & .643 \\
\textbf{\textsc{word2vec}~\cite{mikolov2013efficient}} & .688 & .671 & .663 & .667 & .703 & .714 & .722 & .718 \\
\textbf{\textsc{Doc2Vec}~\cite{le2014distributed}} & .698 & .684 & .712 & .698 & .615 & .610 & .620 & .615 \\ 
\textbf{Castillo et al.~\cite{castillo2011information}} & .794 & .764 & .889 & .822 & .789 & .815 & .774 & .794 \\
\textbf{Shu et al.~\cite{shu2019beyond}} & .878 & .867 & .893 & .880 & .864 & .849 & .893 & .870 \\ \hline
\textbf{Our Model} & \textbf{.892} & \textbf{.877} & \textbf{.908} & \textbf{.892} & \textbf{.879} & \textbf{.857} & \textbf{.902} & \textbf{.879} \\
\bottomrule[1pt]
\end{tabular}
\end{table}

\footnotetext[10]{For each dataset, the maximum value is underlined, that in each column is bold, and that in each row is colored in gray.}

\subsubsection{Fake News Analysis Across Language Levels} \label{subsubsec:acrossLevel}
As being specified in Section \ref{sec:method}, features representing news content are extracted at lexicon-level, syntax-level, semantic-level and discourse-level. We first evaluate the performance of such features within or across language levels in predicting fake news in (\textbf{E1}), followed by feature importance analysis at each level in (\textbf{E2}). 

\begin{table}[p]
\footnotesize
\caption{Feature Performance across Language Levels\protect\footnotemark[10]. Lexicon-level and deep syntax-level features outperform the others, where the performance of semantic-level and shallow syntax-level ones follows. When combining features (exclude RRs) across levels, it enhances the performance compared to when separately using them in predicting fake news.}
\label{tab:acrossLevel}
\begin{tabular}{crrcccc|cccc}
\toprule[1pt]
 &  &  & \multicolumn{4}{c|}{\textbf{PolitiFact}} & \multicolumn{4}{c}{\textbf{BuzzFeed}} \\ \cline{4-11} 
 &  &  & \multicolumn{2}{c}{\textbf{XGBoost}} & \multicolumn{2}{c|}{\textbf{RF}} & \multicolumn{2}{c}{\textbf{XGBoost}} & \multicolumn{2}{c}{\textbf{RF}} \\ \cline{4-11} 
\multirow{-3}{*}{} & \multirow{-3}{*}{\textbf{Language Level}} & \multirow{-3}{*}{\textbf{Feature Group}} & \textbf{Acc.} & \textbf{$\mathbf{F_1}$} & \textbf{Acc.} & \textbf{$\mathbf{F_1}$} & \textbf{Acc.} & \textbf{$\mathbf{F_1}$} & \textbf{Acc.} & \textbf{$\mathbf{F_1}$} \\ \hline
 & \textbf{Lexicon} & BOW & \cellcolor[gray]{0.9}.856 & \cellcolor[gray]{0.9}.858 & .837 & .836 & \cellcolor[gray]{0.9}.823 & \cellcolor[gray]{0.9}.823 & .815 & .815 \\ 
 & \textbf{Shallow Syntax} & POS & .755 & .755 & \cellcolor[gray]{0.9}.776 & \cellcolor[gray]{0.9}.776 & \cellcolor[gray]{0.9}.745 & \cellcolor[gray]{0.9}.745 & .732 & .732 \\
 & \textbf{Deep Syntax} & CFG & \cellcolor[gray]{0.9}.877 & \cellcolor[gray]{0.9}.877 & .836 & .836 & .778 & .778 & \cellcolor[gray]{0.9}.845 & \cellcolor[gray]{0.9}.845 \\ 
 & \textbf{Semantic} & DIA+CBA & \cellcolor[gray]{0.9}.745 & \cellcolor[gray]{0.9}.748 & 737 & .737 & .722 & .750 & \cellcolor[gray]{0.9}.789 & \cellcolor[gray]{0.9}.789 \\
\multirow{-5}{*}{\textbf{\begin{tabular}[c]{@{}c@{}}Within \\ Levels\end{tabular}}} & \textbf{Discourse} & RR & .621 & .621 & \cellcolor[gray]{0.9}.633 & \cellcolor[gray]{0.9}.633 & .658 & .658 & \cellcolor[gray]{0.9}.665 & \cellcolor[gray]{0.9}.665 \\ \hline
 & \textbf{Lexicon+Syntax} & BOW+POS+CFG & \cellcolor[gray]{0.9}.858 & \cellcolor[gray]{0.9}.860 & .822 & .822 & .845 & .845 & \cellcolor[gray]{0.9}\textbf{.871} & \cellcolor[gray]{0.9}\textbf{.871} \\
 & \textbf{Lexicon+Semantic} & BOW+DIA+CBA & \cellcolor[gray]{0.9}.847 & \cellcolor[gray]{0.9}.820 & .839 & .839 & \cellcolor[gray]{0.9}.844 & \cellcolor[gray]{0.9}.847 & .844 & .844 \\
 & \textbf{Lexicon+Discourse} & BOW+RR & .877 & .877 & \cellcolor[gray]{0.9}.880 & \cellcolor[gray]{0.9}.880 & \cellcolor[gray]{0.9}.872 & \cellcolor[gray]{0.9}.873 & .841 & .841 \\
 & \textbf{Syntax+Semantic} & POS+CFG+DIA+CBA & \cellcolor[gray]{0.9}.879 & \cellcolor[gray]{0.9}.880 & .827 & .827 & .817 & .823 & \cellcolor[gray]{0.9}.844 & \cellcolor[gray]{0.9}.844 \\
 & \textbf{Syntax+Discourse} & POS+CFG+RR & \cellcolor[gray]{0.9}.858 & \cellcolor[gray]{0.9}.858 & .813 & .813 & .817 & .823 & \cellcolor[gray]{0.9}.844 & \cellcolor[gray]{0.9}.844 \\
\multirow{-6}{*}{\textbf{\begin{tabular}[c]{@{}c@{}}Across\\ Two\\ Levels\end{tabular}}} & \textbf{Semantic+Discourse} & DIA+CBA+RR & .855 & .857 & \cellcolor[gray]{0.9}.864 & \cellcolor[gray]{0.9}.864 & .844 & .841 & \cellcolor[gray]{0.9}.847 & \cellcolor[gray]{0.9}.847 \\ \hline
 & \textbf{All-Lexicon} & All-BOW & .870 & .870 & \cellcolor[gray]{0.9}.871 & \cellcolor[gray]{0.9}.871 & .851 & .844 & \cellcolor[gray]{0.9}.856 & \cellcolor[gray]{0.9}.856 \\
 & \textbf{All-Syntax} & All-POS-CFG & \cellcolor[gray]{0.9}.834 & \cellcolor[gray]{0.9}.834 & .822 & .822 & \cellcolor[gray]{0.9}.844 & \cellcolor[gray]{0.9}.844 & .822 & .822 \\
 & \textbf{All-Semantic} & All-DIA-CBA & \cellcolor[gray]{0.9}.868 & \cellcolor[gray]{0.9}.868 & .852 & .852 & .848 & .847 & \cellcolor[gray]{0.9}.866 & \cellcolor[gray]{0.9}.866 \\
\multirow{-4}{*}{\textbf{\begin{tabular}[c]{@{}c@{}}Across\\ Three\\ Levels\end{tabular}}} & \textbf{All-Discourse} & All-RR & \cellcolor[gray]{0.9}{\underline{\textbf{.892}}} & \cellcolor[gray]{0.9}{\underline{\textbf{.892}}} & \textbf{.887} & \textbf{.887} & \cellcolor[gray]{0.9}{\underline{\textbf{.879}}} & \cellcolor[gray]{0.9}{\underline{\textbf{.879}}} & .868 & .868 \\ \hline
\multicolumn{3}{r}{{Overall}} & \cellcolor[gray]{0.9}.865 & \cellcolor[gray]{0.9}.865 & .845 & .845 & \cellcolor[gray]{0.9}.855 & \cellcolor[gray]{0.9}.856 & .854 & .854 \\  \bottomrule[1pt]
\end{tabular}
\end{table}

\begin{table}[p]
\vspace{1em}
\small
\caption{Important Lexicon-level, Syntax-level and Discourse-level Features for Fake News Detection.}
\begin{center}
\label{tab:important_features}
\begin{minipage}{0.4\textwidth}
\centering
\subtable[Lexicons]{ \label{subtab:bow}
\begin{tabular}{ccc}
\toprule[1pt]
\textbf{Rank} & \textbf{PolitiFact} & \textbf{BuzzFeed} \\ \hline
1 & `nominee' & `said' \\
2 & `continued' & `authors' \\
3 & `story' & `university' \\
4 & `authors' & `monday' \\
5 & `hillary' & `one' \\
6 & `presidential' & `trump' \\
7 & `highlight' & `york' \\
8 & `debate' & `daily' \\
9 & `cnn' & `read' \\
10 & `republican' & `donald' \\ \bottomrule[1pt]
\end{tabular}} 
\subtable[POS Tags]{ \label{subtab:POS}
\begin{tabular}{ccc}
\toprule[1pt]
\textbf{Rank} & \textbf{PolitiFact} & \textbf{BuzzFeed} \\ \hline
1 & POS & NN \\
2 & JJ & VBN \\
3 & VBN & POS \\
4 & IN & JJ \\
5 & VBD & RB \\ \bottomrule
\end{tabular}}
\end{minipage} 
\begin{minipage}{0.55\textwidth}
\centering
\subtable[Rewrite Rules]{ \label{subtab:cfg}
\begin{tabular}{ccc}
\toprule[1pt]
\textbf{Rank} & \textbf{PolitiFact} & \textbf{BuzzFeed} \\ \hline
1 & NN $\rightarrow$ `story' & VBD $\rightarrow$ `said' \\
2 & NP $\rightarrow$ NP NN & ADVP $\rightarrow$ RB NP \\
3 & VBD $\rightarrow$ `said' & RB $\rightarrow$ `hillary' \\
4 & ROOT $\rightarrow$ S & NN $\rightarrow$ `university' \\
5 & POS $\rightarrow$ `'s' & NNP $\rightarrow$ `monday' \\
6 & NN $\rightarrow$ 'republican' & VP $\rightarrow$ VBD NP NP \\
7 & NN $\rightarrow$ `york' & NP $\rightarrow$ NNP \\
8 & NN $\rightarrow$ `nominee' & VP $\rightarrow$ VB NP ADVP \\
9 & JJ $\rightarrow$ `hillary' & S $\rightarrow$ ADVP VP \\
10 & JJ $\rightarrow$ `presidential' & NP $\rightarrow$ JJ \\
\bottomrule[1pt]
\end{tabular}} 
\subtable[RRs]{ \label{subtab:rr}
\begin{tabular}{ccc}
\toprule[1pt]
\textbf{Rank} & \textbf{PolitiFact} & \textbf{BuzzFeed} \\ \hline
1 & nucleus & attribution \\
2 & attribution & nucleus \\
3 & textualorganization & satellite \\
4 & elaboration & span \\
5 & same\_unit & same\_unit \\
\bottomrule[1pt]
\end{tabular}}
\end{minipage}
\end{center}
\end{table}

\vspace{0.5em}
\noindent \textbf{E1:} \textit{Feature Performance Across Language Levels.} Table \ref{tab:acrossLevel} presents the performance of features within each level and across levels for fake news detection. Results indicate that within single level, (1)~features at lexicon-level (BOWs) and deep syntax-level (CFGs) outperform the others, which can achieve above 80\% accuracy rate and $F_1$ score, where (2) the performance of features at semantic-level (DIAs and CBAs) and shallow syntax-level (POS tags) follows with an accuracy and $F_1$ score that is between 70\% to 80\%. However, (3) fake news prediction using the standardized frequencies of rhetorical relationships (discourse-level) does not perform well within the framework. It should be noted that the number of features based on BOWs and CFGs is in the order of a thousand, much more than others that are within the order of a hundred; and (4) when combining features (exclude RRs) across levels, it enhances the performance compared to when separately using features within each level in predicting fake news. Such performance can achieve an accuracy value and $F_1$ score around $\sim$88\%. In addition, it can be observed from Table \ref{tab:generalPerformance} and Table \ref{tab:acrossLevel} that though the assessment of semantic-level features (DIAs and CBAs) that we defined and selected based on psychological theories rely on LIWC, their performance in predicting fake news is better than directly utilizing all word and phrase categories provided by LIWC without supportive theories.

\vspace{0.5em}
\noindent \textbf{E2:} \textit{Feature Importance Analysis.} RF (mean decrease impurity) is used to determine the importance of features, among which the top discriminating lexicons, POS tags, rewrite rules and RRs are provided in
Table \ref{tab:important_features}.
It can be seen that (1) discriminating lexicons differ from one dataset to the other; (2) compared to the other POS tags, the standardized frequencies of POS (possessive ending), VBN (verb in a form of past participle) and JJ (adjective) can better differentiate fake news from true news in two datasets; (3) unsurprisingly, discriminating rewrite rules are often formed based on discriminating lexicons and POS tags, e.g., JJ $\rightarrow$ `presidential' and ADVP (adverb phrase) $\rightarrow$ RB (adverb) NP (noun phrase); (4) compared to the other RRs, 
nucleus that contains basic information about parts of text and 
same\_unit that indicates the relation between discontinuous clauses play a comparatively significant role in predicting fake news. It should be noted that though these features can capture news content style and perform well, they are not as easy to understand as semantic-level features. Considering that, detailed analyses for DIAs (Section \ref{subsubsec:deception}) and CBAs (Section \ref{subsubsec:clickbait}) are conducted next.

\subsubsection{Types of Deception and Fake News} \label{subsubsec:deception}
As discussed in Section \ref{subsubsec:semantic}, well-established forensic psychology theories on identifying deception/disinformation have inspired us to represent news content by measuring its [psycho-linguistic] attributes, e.g., sentiment. Such potential clues provided by these theories help reveal fake news patterns that are easy to understand. Opportunities are also provided to compare types of deception/disinformation and fake news; theoretically, deception/disinformation is a more general concept compared to fake news, which additionally includes fake statements, fake reviews, and the like. 
In this section, we first evaluate the performance of these disinformation-related attributes (i.e., DIAs) in predicting fake news in (\textbf{E1}). Then in (\textbf{E2}), important features and attributes are identified, followed by a detailed feature analysis to reveal the potential patterns of fake news and compare them with that of deception (\textbf{E3}).

\vspace{0.5em}
\noindent \textbf{E1:} \textit{Performance of Disinformation-related Attributes in Predicting Fake News.} Table \ref{tab:dia} presents the performance of disinformation-related attributes in predicting fake news. Results indicate that identifying fake news articles respectively based on their content quality, sentiment, quantity, and specificity performs similarly, with 60\% to 70\% accuracy and $F_1$ score using PolitiFact data, and 50\% to 60\% accuracy and $F_1$ score using BuzzFeed data.
Combining all attributes to detect fake news performs better than separately using each type of attribute, which can achieve 70\% to 80\% accuracy and $F_1$ score on PolitiFact data, and 60\% to 70\% accuracy and $F_1$ score on BuzzFeed data.

\begin{table}[t]
\small
\caption{Performance of Disinformation-related Attributes in Predicting Fake News\protect\footnotemark[10]. Individual attributes perform similarly while combining all attributes perform better in predicting fake news.}
\label{tab:dia}
\begin{tabular}{lcccc|cccc}
\toprule[1pt]
 & \multicolumn{4}{c|}{\textbf{PolitiFact}} & \multicolumn{4}{c}{\textbf{BuzzFeed}} \\ \cline{2-9} 
 & \multicolumn{2}{c}{\textbf{XGBoost}} & \multicolumn{2}{c|}{\textbf{RF}} & \multicolumn{2}{c}{\textbf{XGBoost}} & \multicolumn{2}{c}{\textbf{RF}} \\ \cline{2-9} 
\multirow{-3}{*}{\textbf{\begin{tabular}[c]{@{}l@{}}Disinformation-\\ related Attribute(s)\end{tabular}}} & \textbf{Acc.} & $\mathbf{F_1}$ & \textbf{Acc.} & $\mathbf{F_1}$ & \textbf{Acc.} & $\mathbf{F_1}$ & \textbf{Acc.} & $\mathbf{F_1}$ \\ \hline
\textbf{Quality} & \cellcolor[gray]{0.9}.667 & \cellcolor[gray]{0.9}.652 & .645 & .645 & \cellcolor[gray]{0.9}.556 & \cellcolor[gray]{0.9}.500 & .512 & .512 \\
\hspace{5mm}-- Informality & \cellcolor[gray]{0.9}.688 & \cellcolor[gray]{0.9}.727 & .604 & .604 & \cellcolor[gray]{0.9}.555 & \cellcolor[gray]{0.9}.513 & .508 & .508 \\
\hspace{5mm}-- Subjectivity & \cellcolor[gray]{0.9}.688 & \cellcolor[gray]{0.9}.706 & .654 & .654 & \cellcolor[gray]{0.9}.611 & \cellcolor[gray]{0.9}.588 & .533 & .530 \\
\hspace{5mm}-- Diversity & .583 & .600 & \cellcolor[gray]{0.9}.620 & \cellcolor[gray]{0.9}.620 & \cellcolor[gray]{0.9}.639 & \cellcolor[gray]{0.9}.552 & .544 & .544 \\
\textbf{Sentiment} & \cellcolor[gray]{0.9}.625 & \cellcolor[gray]{0.9}.591 & .583 & .583 & \cellcolor[gray]{0.9}.556 & \cellcolor[gray]{0.9}.579 & .515 & .525 \\
\textbf{Quantity} & .583 & .524 & \cellcolor[gray]{0.9}.638 & \cellcolor[gray]{0.9}.638 & .528 & .514 & \cellcolor[gray]{0.9}.584 & \cellcolor[gray]{0.9}.586 \\
\textbf{Specificity} & \cellcolor[gray]{0.9}.625 & \cellcolor[gray]{0.9}.609 & .558 & .558 & .583 & .571 & \cellcolor[gray]{0.9}.611 & \cellcolor[gray]{0.9}.611 \\
\hspace{5mm}-- Cognitive Process & \cellcolor[gray]{0.9}.604 & \cellcolor[gray]{0.9}.612 & .565 & .565 & \cellcolor[gray]{0.9}.556 & \cellcolor[gray]{0.9}.579 & .531 & .531 \\
\hspace{5mm}-- Perceptual Process & .563 & .571 & \cellcolor[gray]{0.9}.612 & \cellcolor[gray]{0.9}.612 & .556 & .600 & \cellcolor[gray]{0.9}.571 & \cellcolor[gray]{0.9}.571 \\ \midrule[1pt]
\textbf{Overall} & \textbf{.729} & \textbf{.735} & \cellcolor[gray]{0.9}\textbf{\underline{.755}} & \cellcolor[gray]{0.9}\textbf{\underline{.755}} & \cellcolor[gray]{0.9}\textbf{\underline{.667}} & \cellcolor[gray]{0.9}\textbf{\underline{.647}} & \textbf{.625} & \textbf{.625} \\ \bottomrule[1pt]
\end{tabular}
\end{table}

\begin{table}[t]
\small
\caption{Important Disinformation-related Features and Attributes for Fake News Detection. In both datasets, content diversity and quantity are most significant in differentiating fake news from the truth; cognitive process involved and content subjectivity are second; content informality and sentiments expressed are third.}
\label{tab:pdaImportance}
\begin{tabular}{cll|ll}
\toprule[1pt]
\multirow{2}{*}{\textbf{Rank}} & \multicolumn{2}{c|}{\textbf{PolitiFact}} & \multicolumn{2}{c}{\textbf{BuzzFeed}} \\ \cline{2-5}
 & \textbf{Feature} & \textbf{Attribute} & \textbf{Feature} & \textbf{Attribute} \\ \hline
1 & \# Characters per Word & Quantity & \# Overall Informal Words & Informality \\
2 & \# Sentences per Paragraph & Quantity & \% Unique Words & Diversity \\
3 & \% Positive Words & Sentiment & \% Unique Nouns & Diversity \\
4 & \% Unique Words & Diversity & \% Unique Content Words & Diversity \\
5 & \% Causation & Cognitive Process & \# Report Verbs & Subjectivity \\
6 & \# Words per Sentence & Quantity & \% Insight & Cognitive Process \\
7 & \% Report Verbs & Subjectivity & \% Netspeak & Informality \\
8 & \% Unique Verbs & Diversity & \# Sentences & Quantity \\
9 & \# Sentences & Quantity & \% Unique Verbs & Diversity \\
10 & \% Certainty Words & Cognitive Process & \% Unique Adverbs & Diversity \\
\bottomrule[1pt]
\end{tabular}
\end{table}

\begin{figure}[t]
    \centering
    \subfigure[PolitiFact]{
    \begin{minipage}{.495\textwidth}
    \includegraphics[width=.495\textwidth]{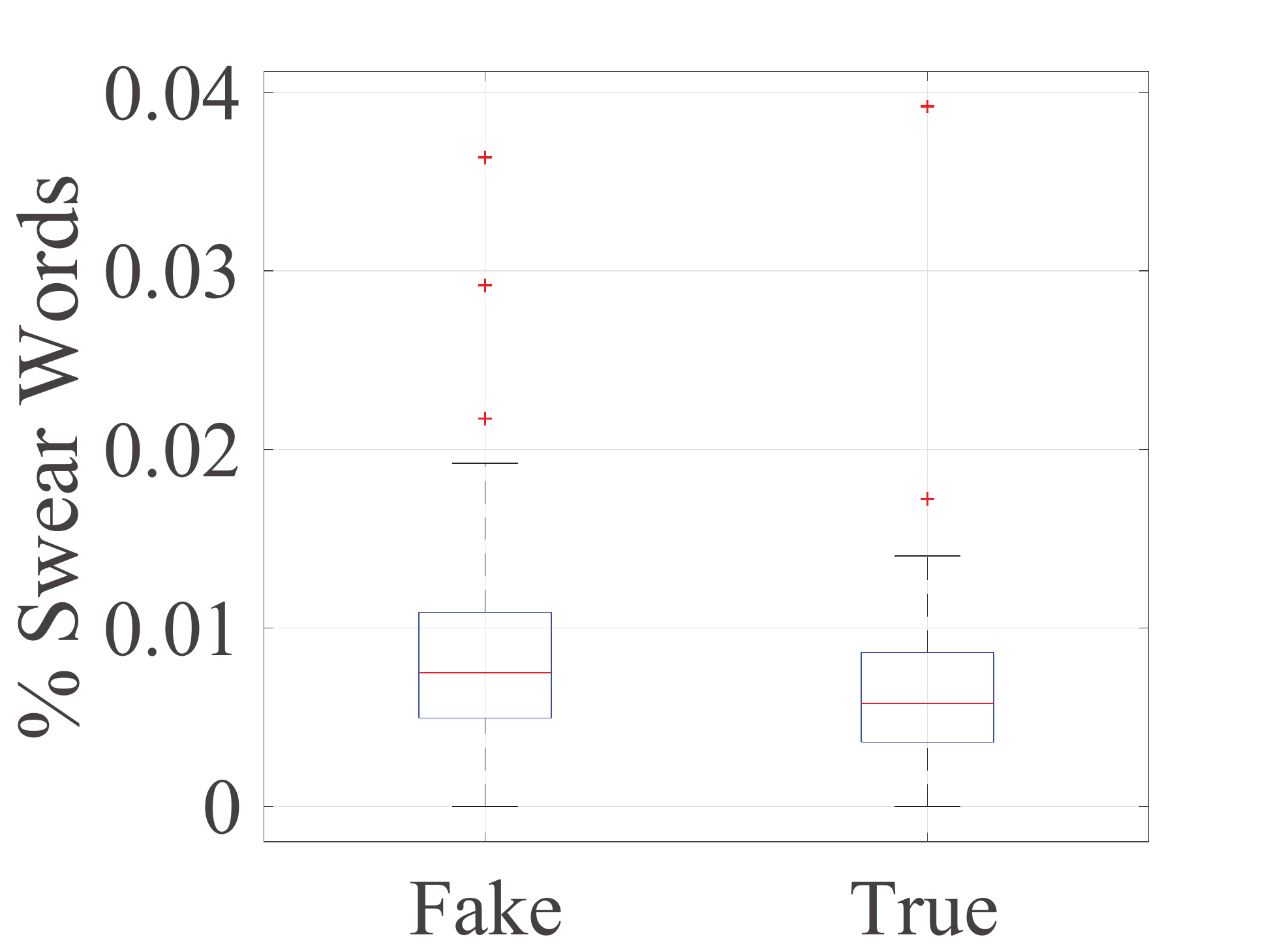}
    \includegraphics[width=.495\textwidth]{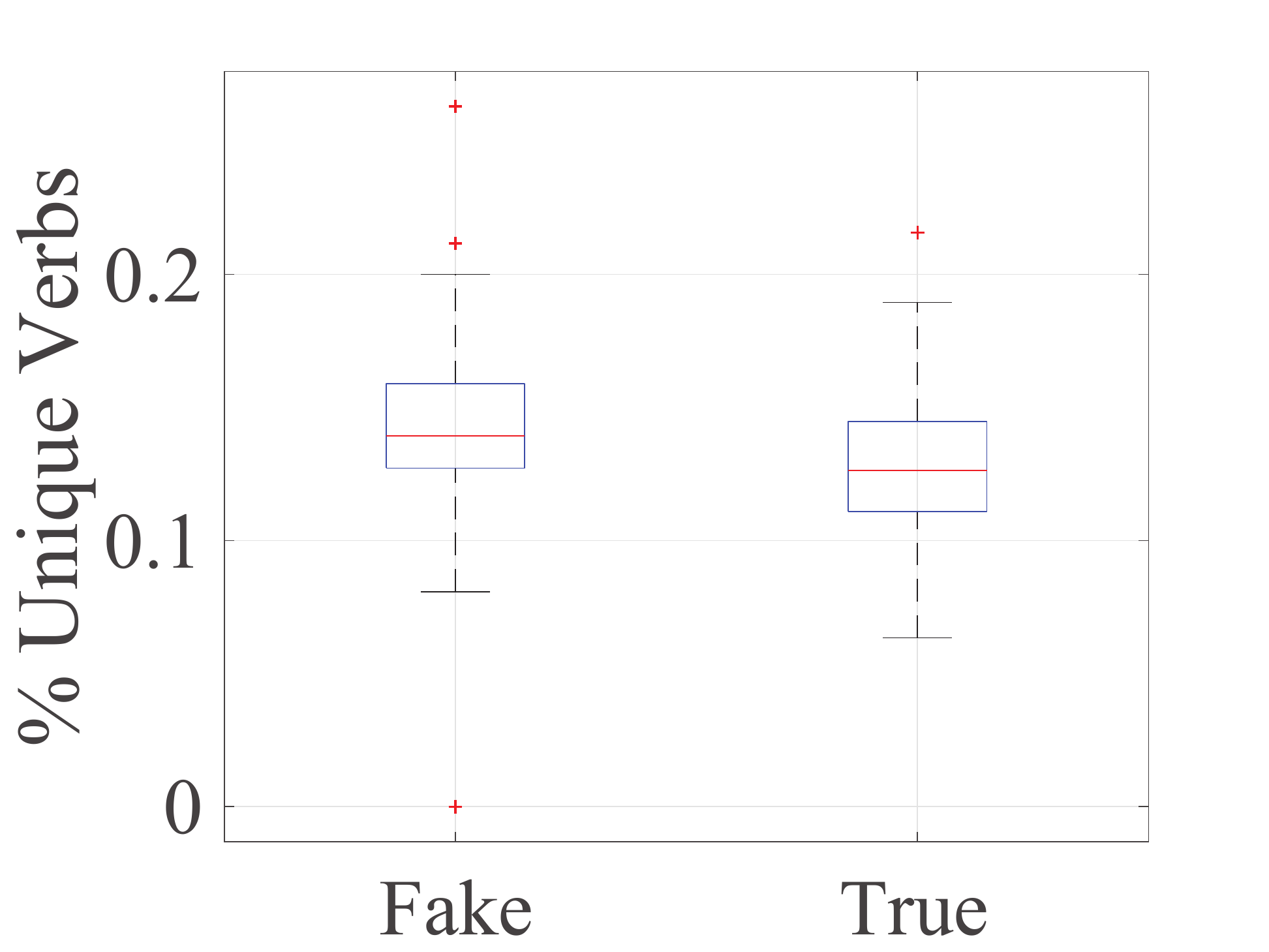}
    \includegraphics[width=.495\textwidth]{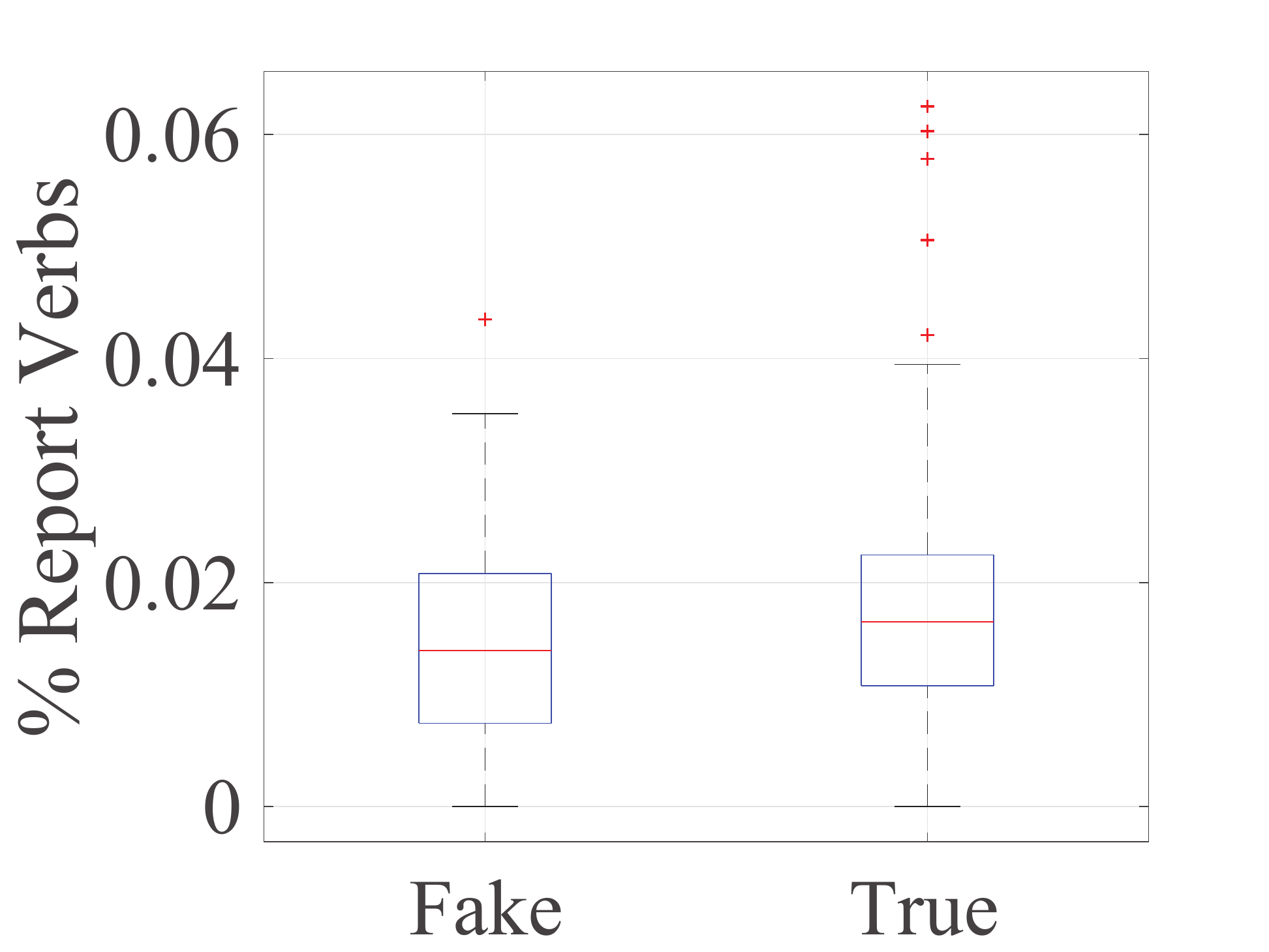}
    \includegraphics[width=.495\textwidth]{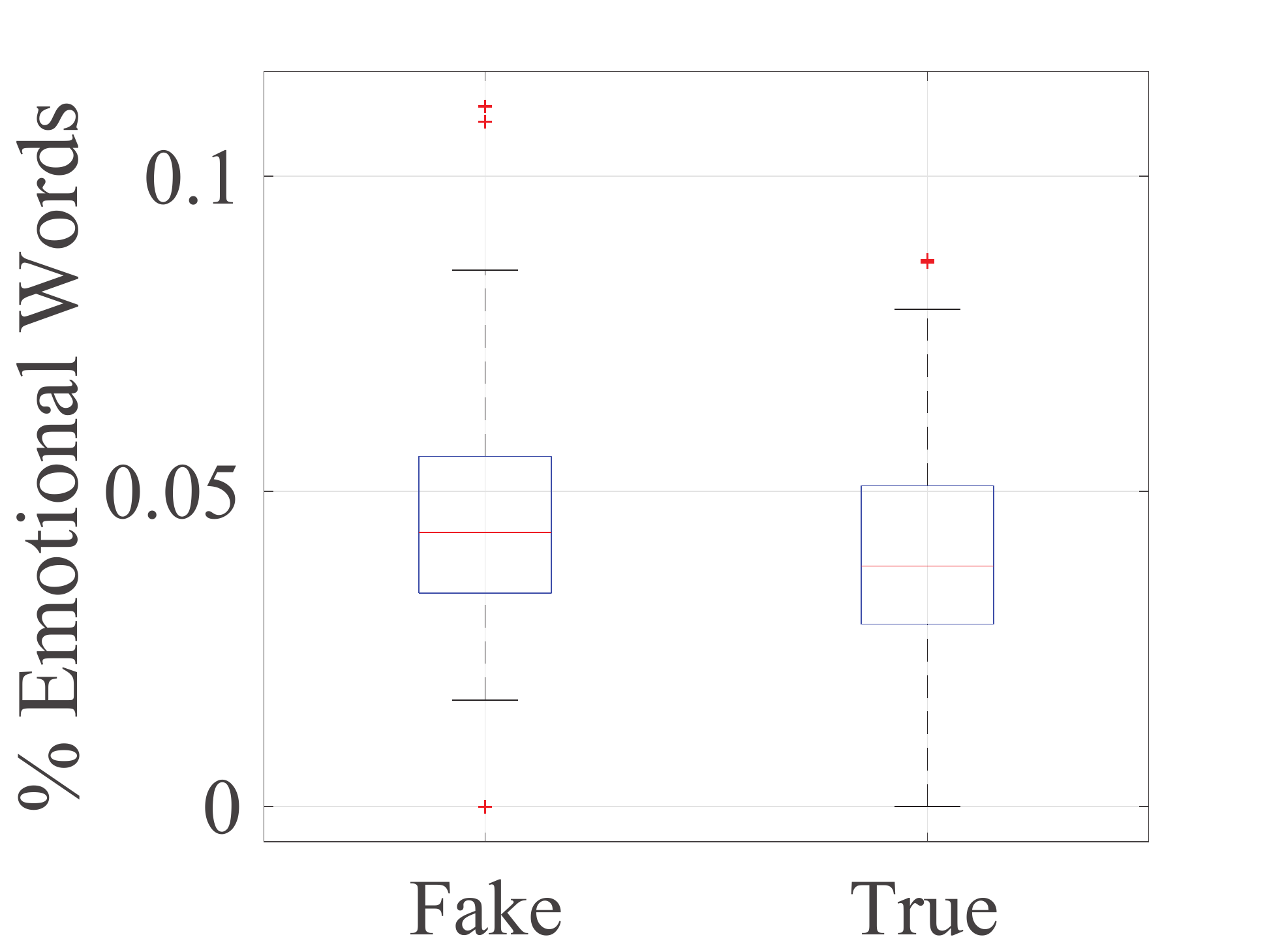}
    \includegraphics[width=.495\textwidth]{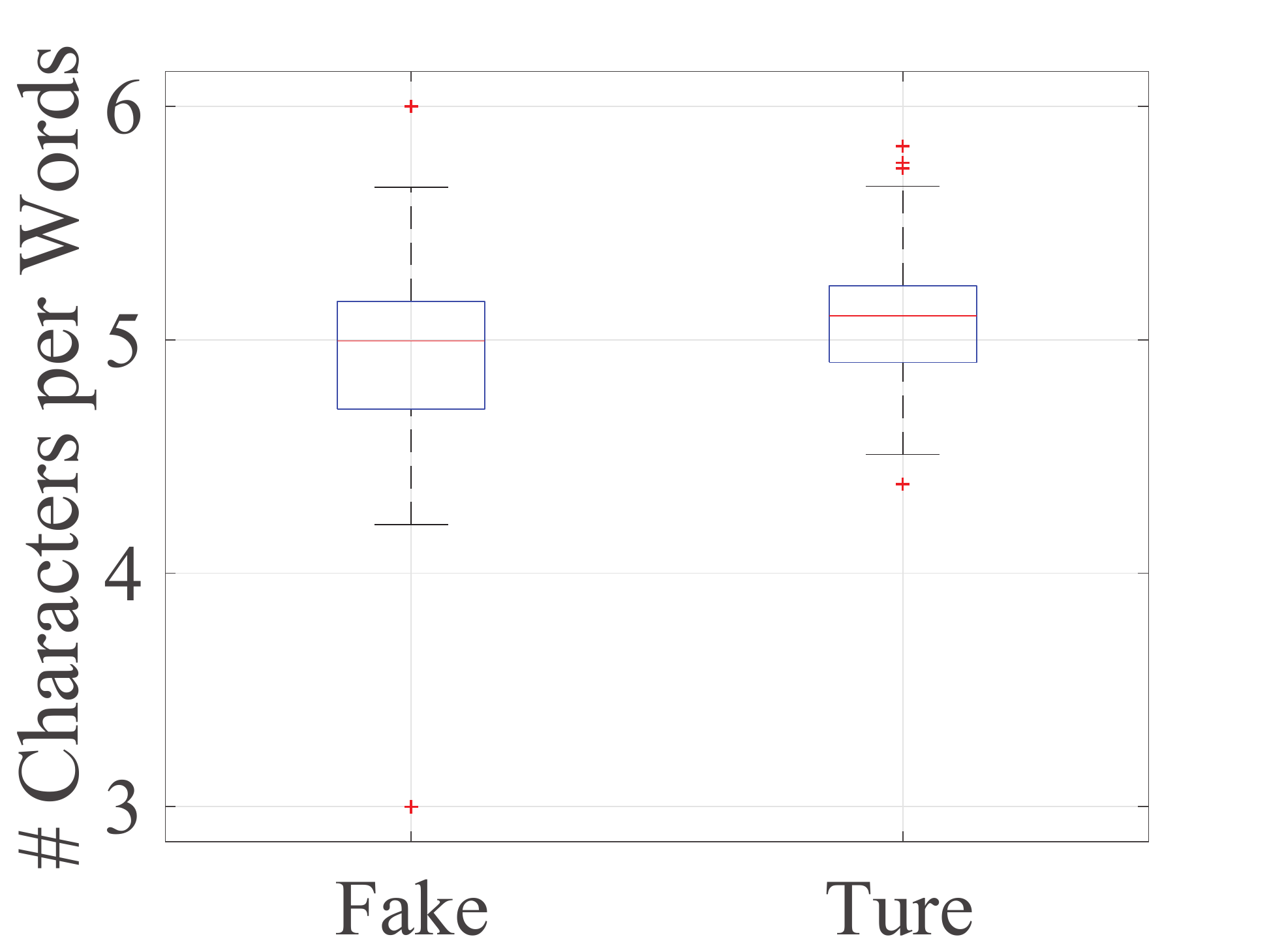}
    \includegraphics[width=.495\textwidth]{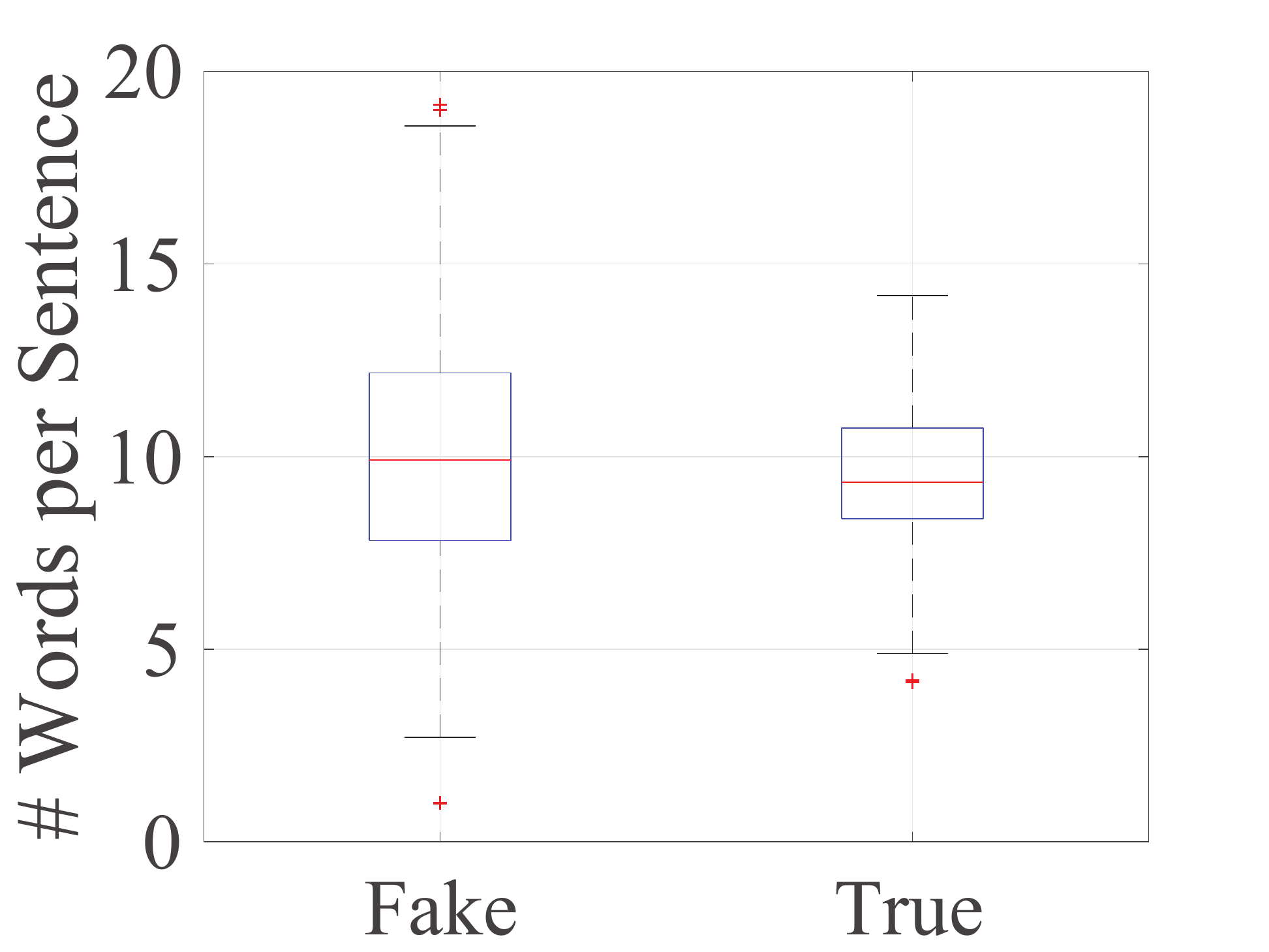}
    \end{minipage}}~
    \subfigure[BuzzFeed]{
    \begin{minipage}{.495\textwidth}
    \includegraphics[width=.495\textwidth]{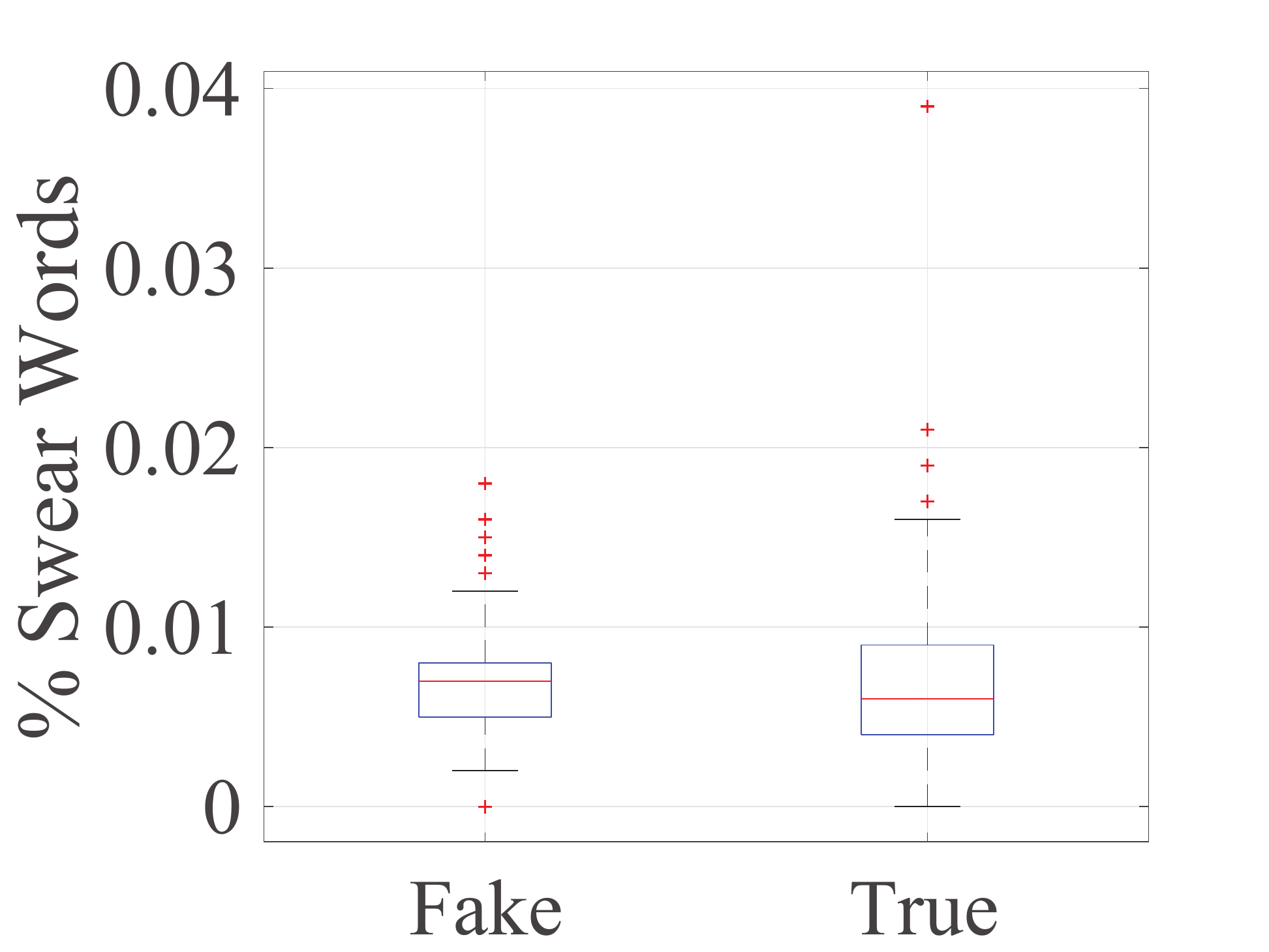}
    \includegraphics[width=.495\textwidth]{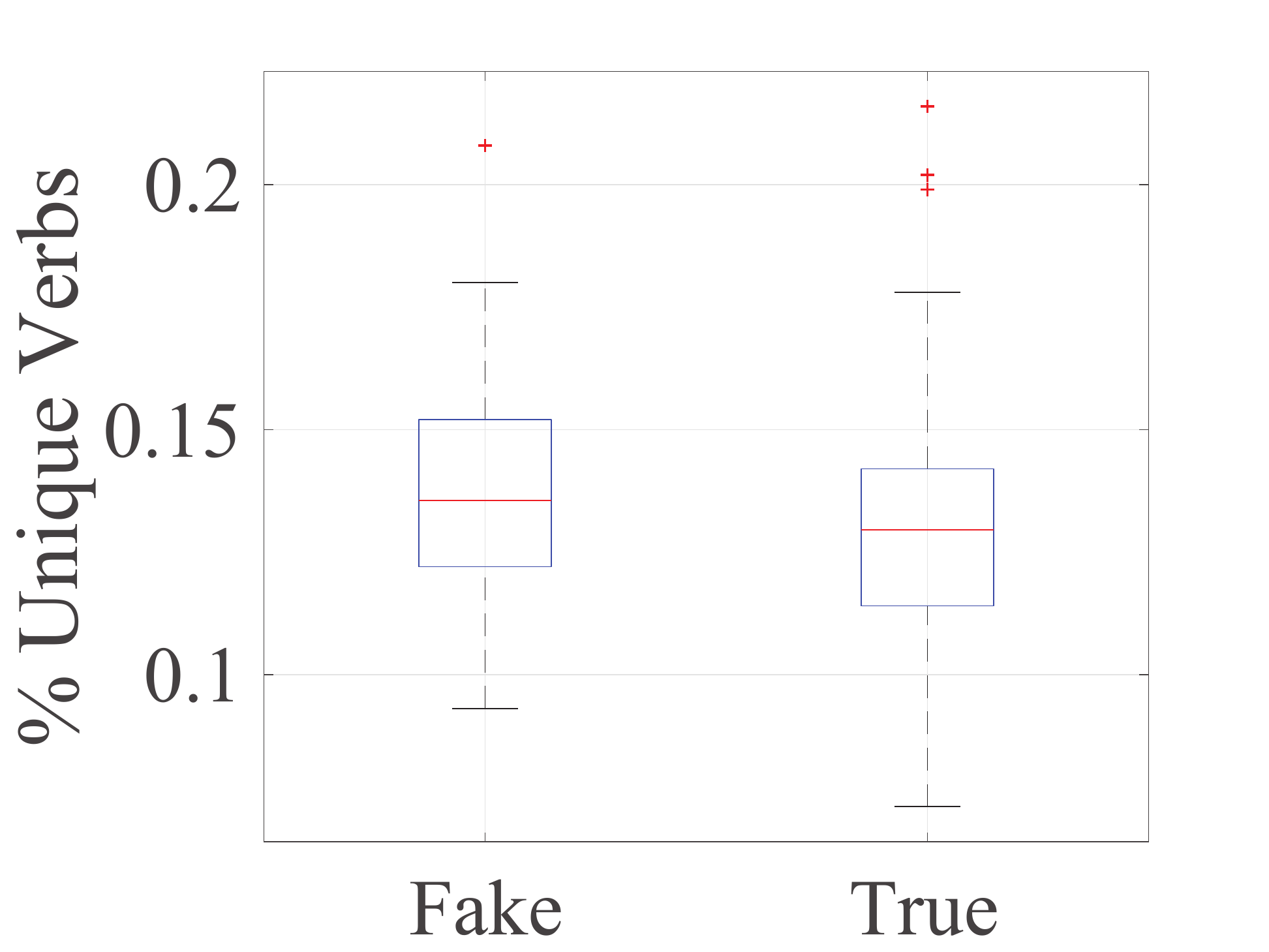}
    \includegraphics[width=.495\textwidth]{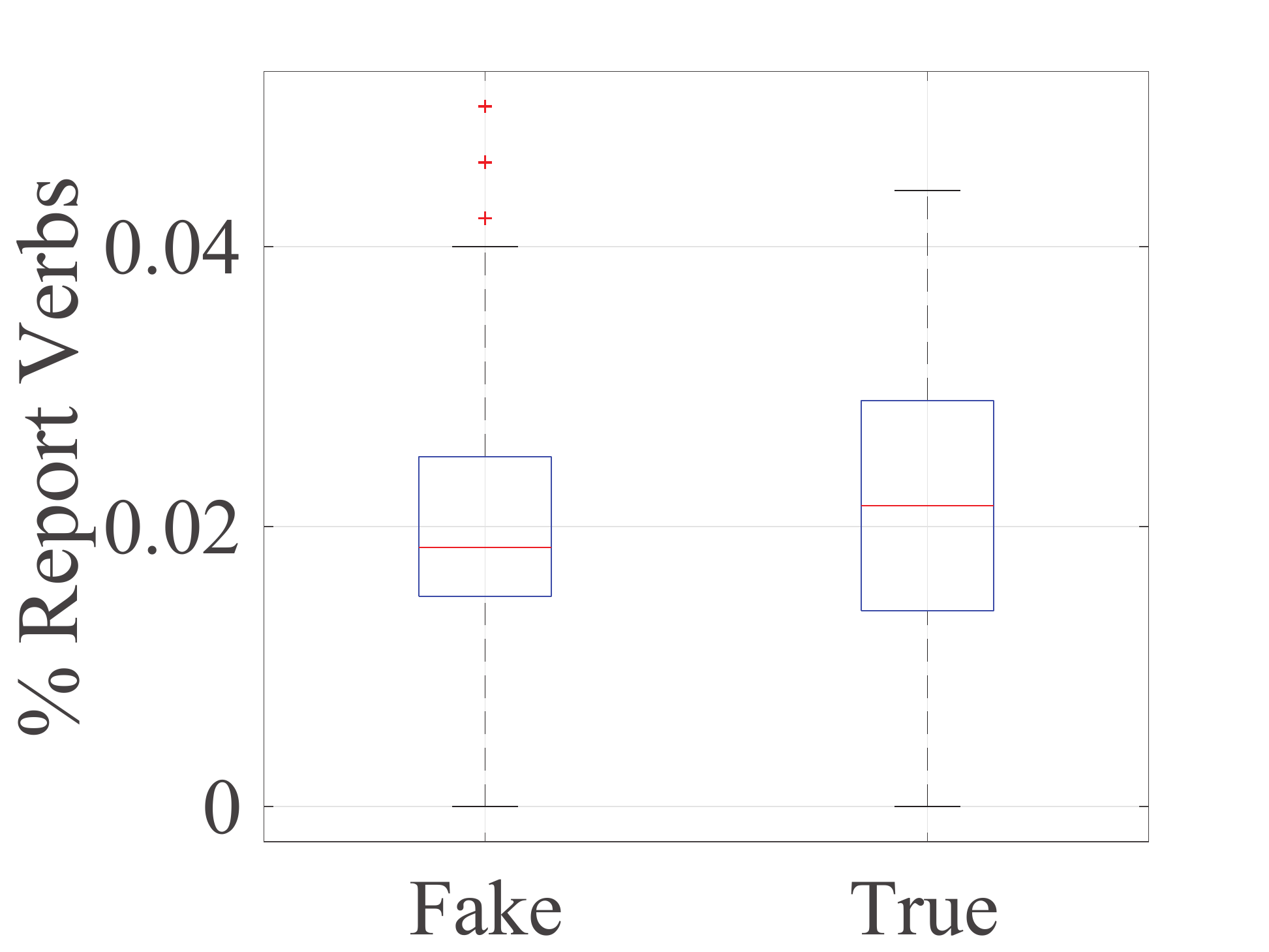}
    \includegraphics[width=.495\textwidth]{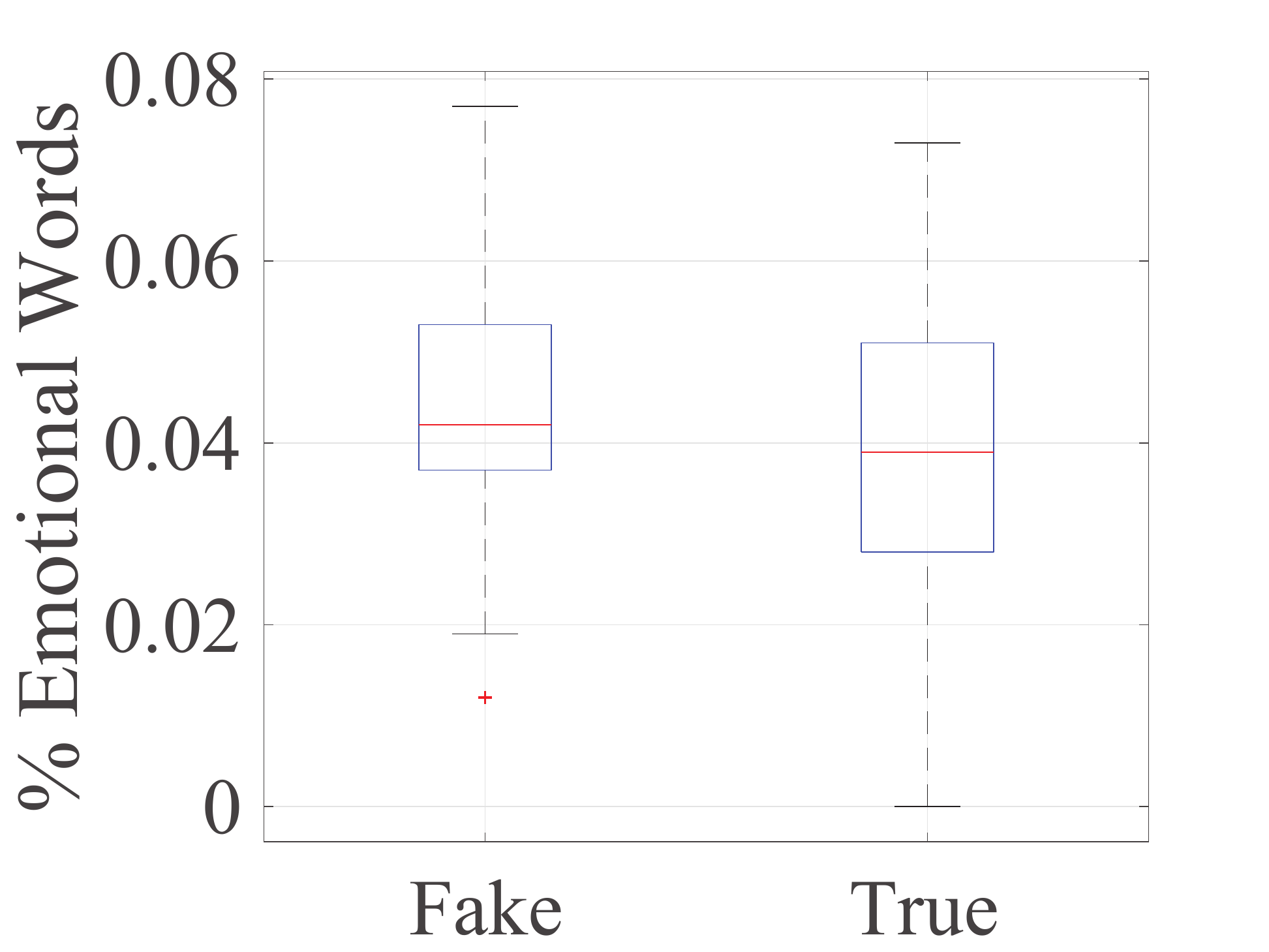}
    \includegraphics[width=.495\textwidth]{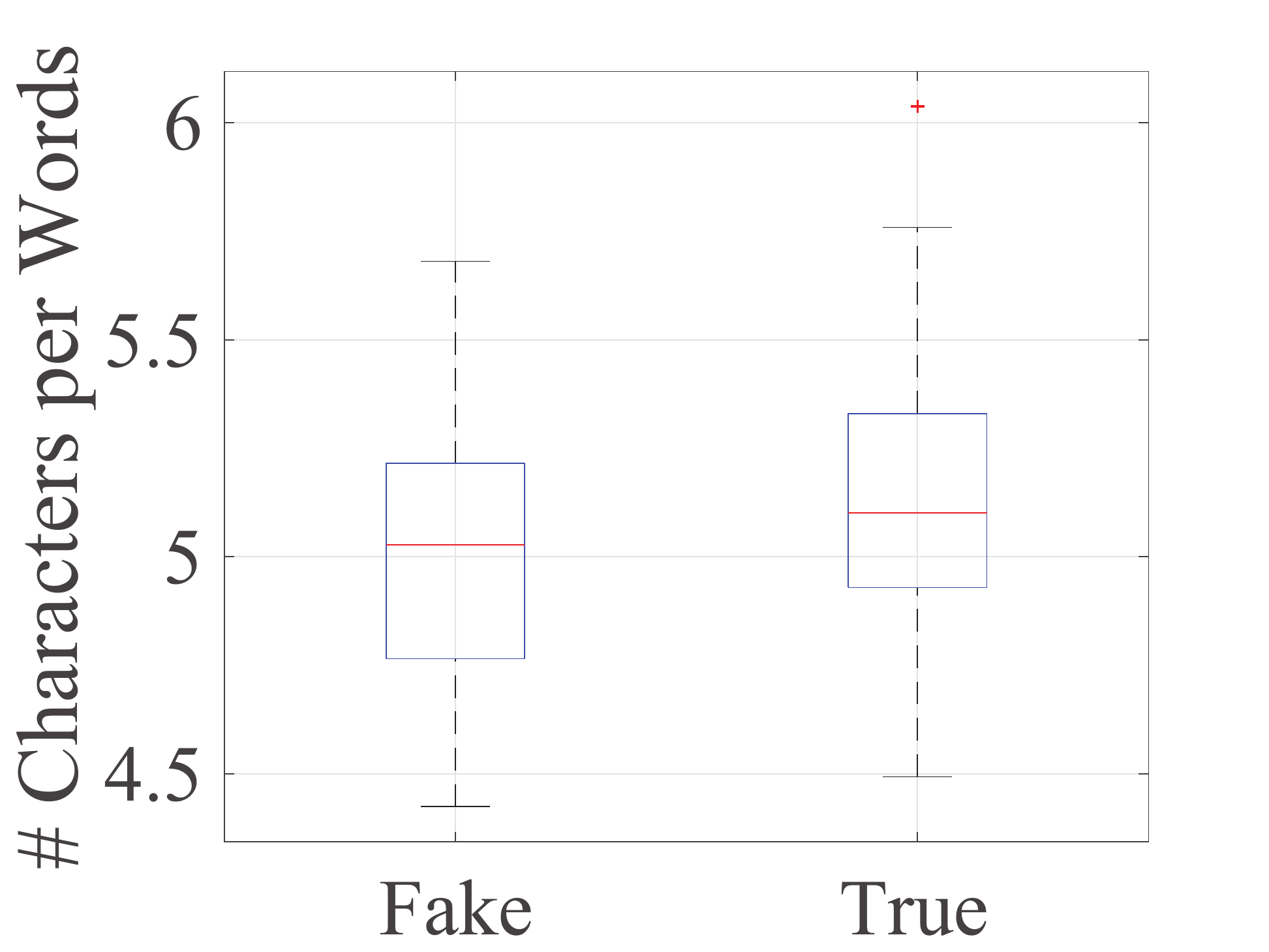}
    \includegraphics[width=.495\textwidth]{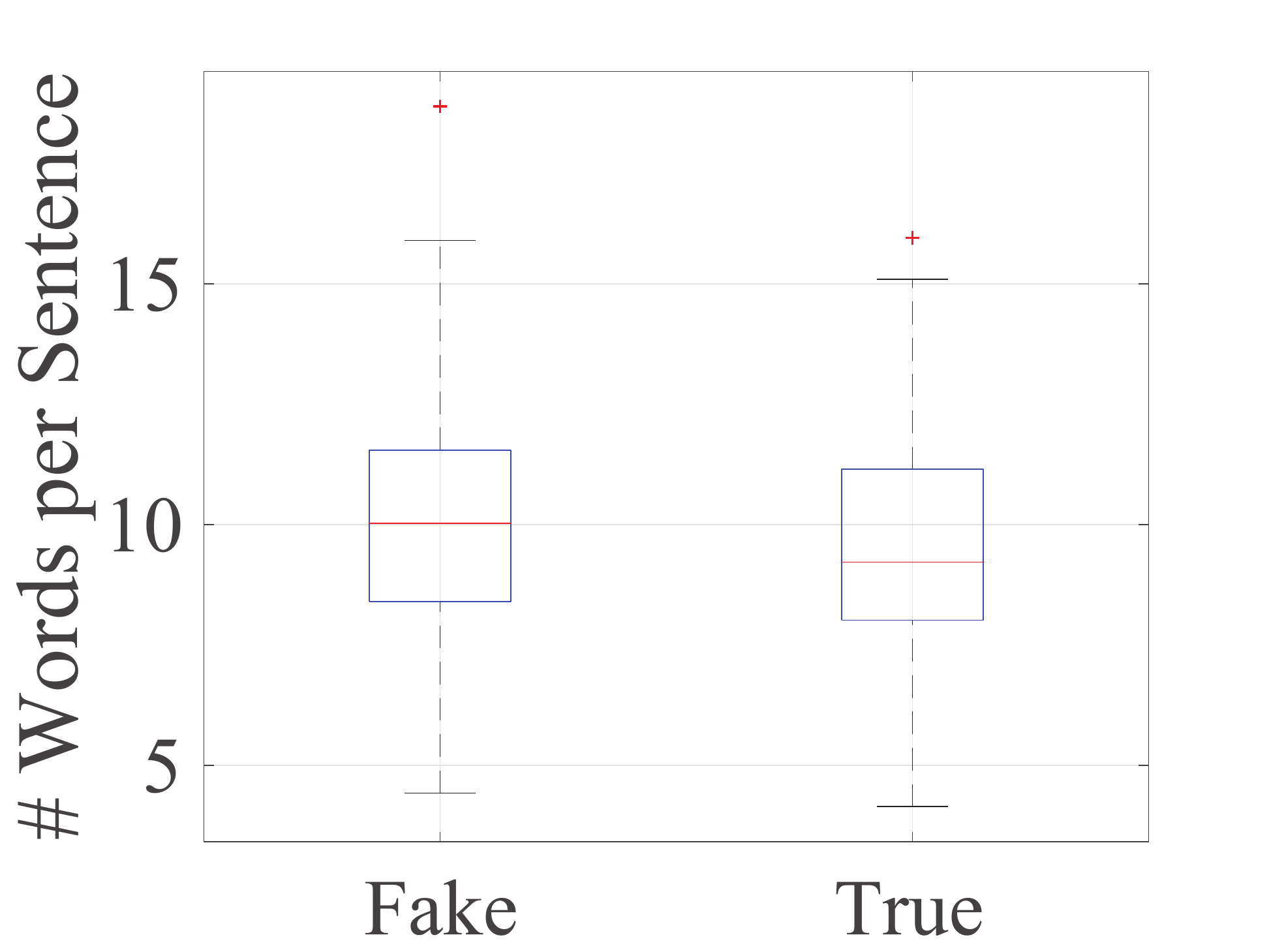}
    \end{minipage}}
    \vspace{-4mm}
    \caption{Potential Patterns of Fake News. In both datasets, fake news shares higher (i) informality (\% swear words), (ii) diversity (\% unique verbs) and (iii) subjectivity (\% report verbs), and is (iv) more emotional (\% emotional words) with (v) longer sentences (\# words per sentence) and (vi) shorter words (\# characters per words) compared to true news. }
    \label{fig:dia_pattern}
\end{figure}

\vspace{0.5em}
\noindent \textbf{E2:} \textit{Importance Analysis for Disinformation-related Features and Attributes.}
RF (mean decrease impurity) is used to determine the importance of features, among which the top ten discriminating features are presented in Table \ref{tab:pdaImportance}. Results indicate that, in general, (1) content quality (i.e., informality, subjectivity and diversity), sentiments expressed, quantity and specificity (i.e., cognitive and perceptual process) all play a role in differentiating fake news articles from the true ones. Specifically, in both datasets, (2) fake news differs more significantly in diversity and quantity from the truth compared to the other attributes, where (3) cognitive process involved in news content and content subjectivity follow. Finally, (4) content informality and sentiments play a comparatively weak role in predicting fake news compared to the others.

\vspace{0.5em}
\noindent \textbf{E3:} \textit{Potential Patterns of Fake News Content.} We analyze each feature in DIA group, among which those that exhibit a consistent pattern in both datasets are presented in Fig. \ref{fig:dia_pattern}.
Specifically, we have the following observations: 

\begin{itemize}
    \item Similar to deception, fake news differs in content quality and sentiments expressed from the truth~\cite{undeutsch1967beurteilung,zuckerman1981verbal}. Compared to true news, fake news often carries less report verbs, while a greater proportion of unique verbs, swear words, and emotional (positive+negative) words. 
    
    \item Compared to true news articles, fake news articles are characterized by shorter words and longer sentences.
    
    \item It is known that deception often does not involve cognitive and perceptual processes~\cite{johnson1981reality,zuckerman1981verbal}. However, the frequencies of lexicons related to cognitive and perceptual processes can hardly discriminate between fake and true news stories based on our datasets. 
\end{itemize}

\subsubsection{Clickbaits and Fake News} \label{subsubsec:clickbait}
We also explore the relationship between clickbaits and fake news by conducting four experiments: (\textbf{E1}) analyzes clickbait distribution within fake and true news articles; (\textbf{E2}) evaluates the performance of clickbait-related attributes in predicting fake news, among which important features and attributes are identified in (\textbf{E3}); and (\textbf{E4}) examines if clickbait and fake news share some potential patterns.

\begin{figure}[t]
    \centering
    \subfigure[PolitiFact]{
    \includegraphics[width=0.32\textwidth]{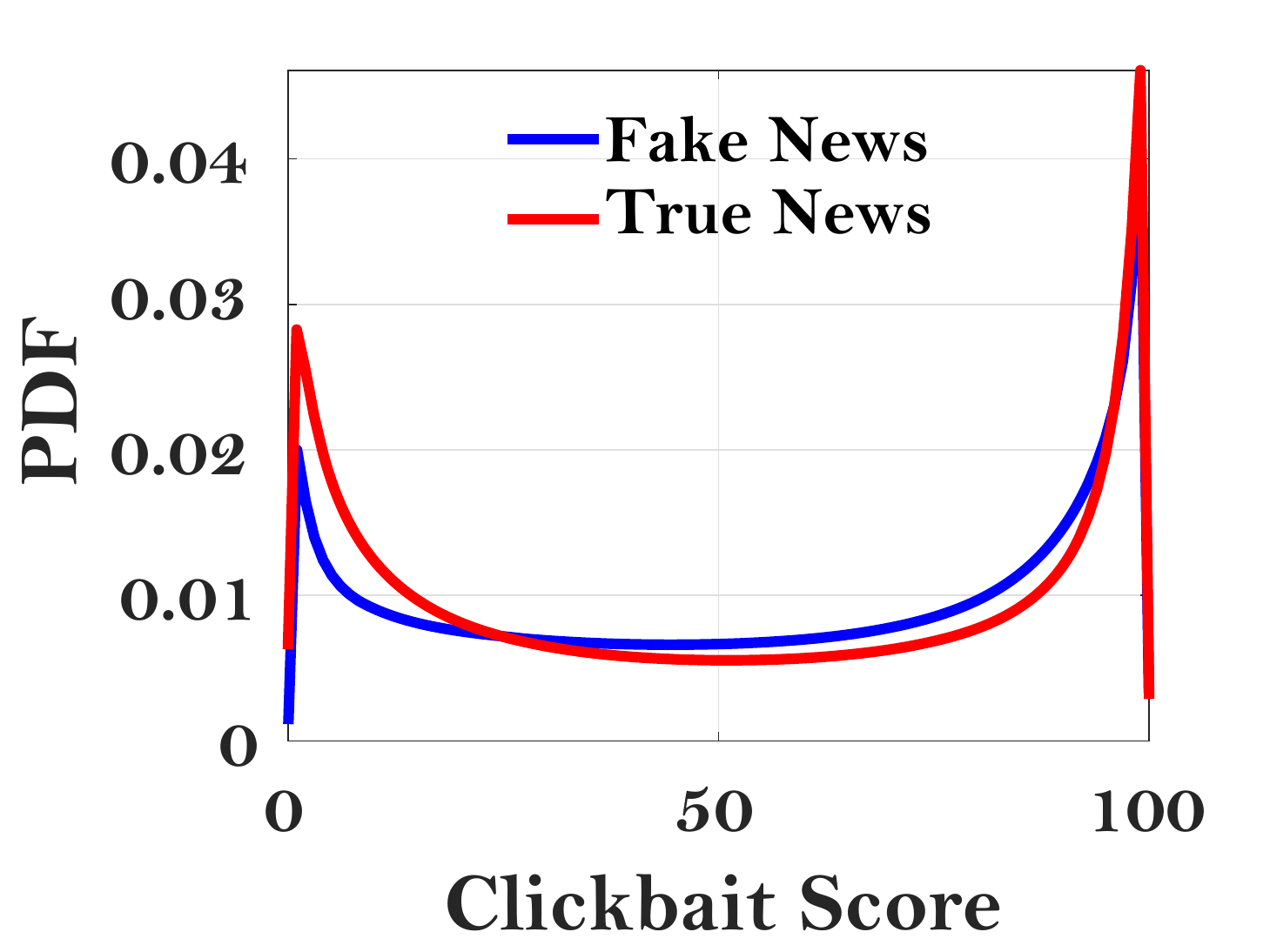}} \qquad
    \subfigure[BuzzFeed]{
    \includegraphics[width=0.32\textwidth]{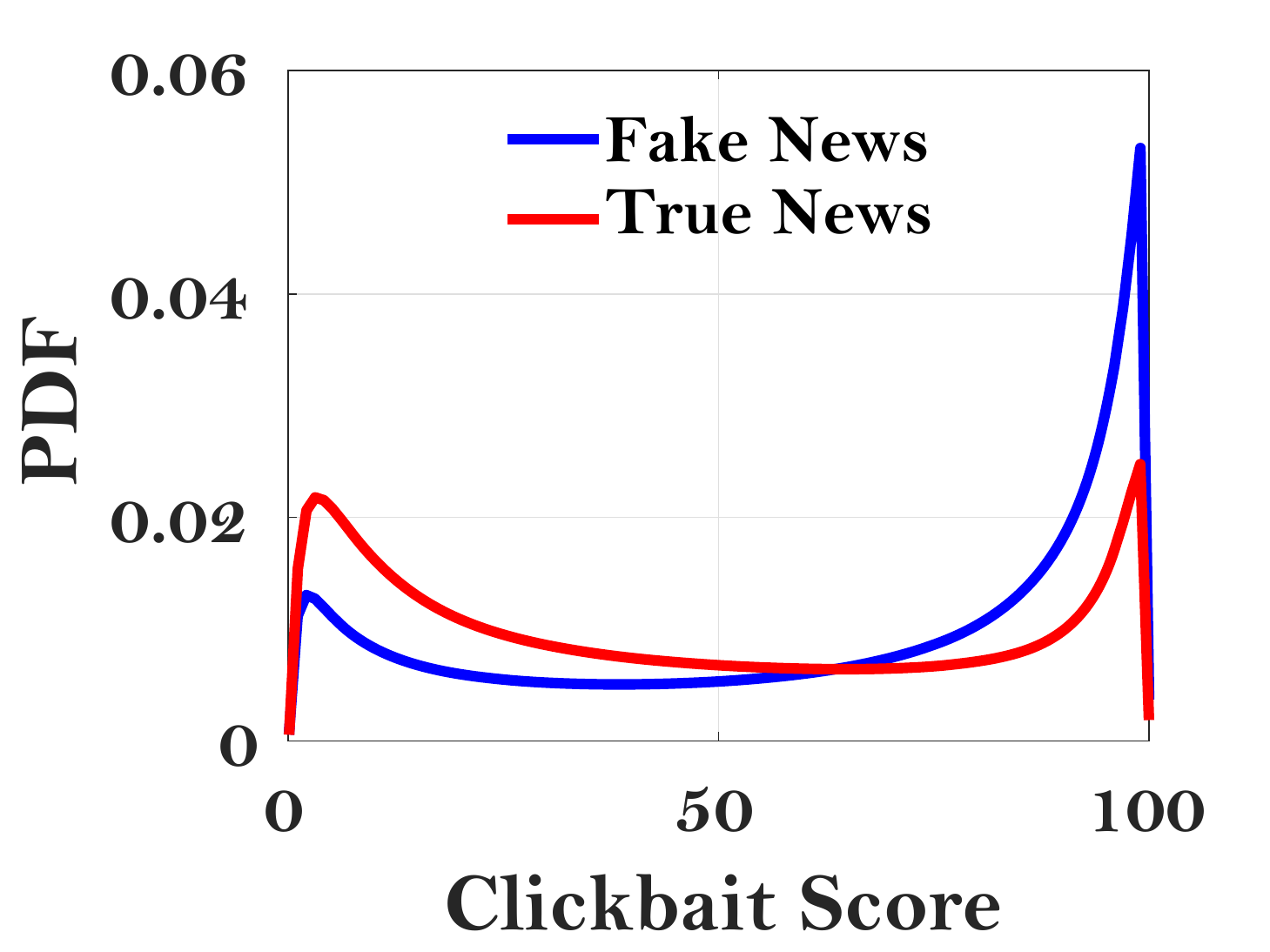}}
    \caption{Clickbait Distribution within Fake and True News Articles. Clickbaits are more common in fake news articles compared to true news articles: among news headlines with relatively high clickbait scores fake news articles often occupy a greater proportion compared to true news articles.}
    \label{fig:clickbait_distribution}
\end{figure}

\vspace{0.5em}
\noindent \textbf{E1:} \textit{Clickbait Distribution within Fake and True News Articles.} As few datasets, including PolitiFact and BuzzFeed, provide both news labels (\textit{fake} or \textit{true}) and news headline labels (\textit{clickbait} or \textit{regular headline}), we use a pretrained deep net, particularly, a Convolutional Neural Network (CNN) model\footnote{\url{https://github.com/saurabhmathur96/clickbait-detector}}~\cite{agrawal2016clickbait} to obtain the clickbait scores ($\in [0,100]$) of news headlines, where 0 indicates not-clickbait (i.e., a regular headline) and 100 indicates clickbait. The model can achieve $\sim$93.8\% accuracy~\cite{agrawal2016clickbait}. Using clickbait scores, we obtain the clickbait distribution (i.e., Probabilistic Density Function, PDF) respectively within fake and true news articles, which is depicted in 
Figure \ref{fig:clickbait_distribution}. We observe that clickbaits have a closer relationship with fake news compared to true news: among news headlines with relatively low clickbait scores, true news articles often occupy a greater proportion compared to fake ones; while among news headlines with relatively high clickbait scores, a greater proportion often refers to fake news articles compared to true news articles.

\vspace{0.5em}
\noindent \textbf{E2:} \textit{Performance of Clickbait-related Attributes in Predicting Fake News.}  Table \ref{tab:clickbait_performance} presents the performance of clickbait-related attributes in predicting fake news. Results indicate that identifying fake news articles based on their headline news-worthiness, whose accuracy and $F_1$ score are around 70\%, performs better than based on either headline readability or sensationalism.

\begin{table}[t]
\small
\caption{Performance of Clickbait-related Attributes in Predicting Fake News\protect\footnotemark[10]. Based on the experimental setup, news-worthiness of headlines outperforms the other attributes in predicting fake news.}
\label{tab:clickbait_performance}
\small
\begin{tabular}{rcccc|cccc}
\toprule[1pt]
 & \multicolumn{4}{c|}{\textbf{PolitiFact}} & \multicolumn{4}{c}{\textbf{BuzzFeed}} \\ \cline{2-9}
 & \multicolumn{2}{c}{\textbf{XGBoost}} & \multicolumn{2}{c|}{\textbf{RF}} & \multicolumn{2}{c}{\textbf{XGBoost}} & \multicolumn{2}{c}{\textbf{RF}} \\ \cline{2-9}
\multirow{-3}{*}{\textbf{\begin{tabular}[c]{@{}r@{}}Clickbait-related \\ Attributes\end{tabular}}} & \textbf{Acc.} & $\mathbf{F_1}$ & \textbf{Acc.} & $\mathbf{F_1}$ & \textbf{Acc.} & $\mathbf{F_1}$ & \textbf{Acc.} & $\mathbf{F_1}$ \\ \hline
\textbf{Readability} & \cellcolor[gray]{0.9}.708 & \cellcolor[gray]{0.9}.682 & .636 & .636 & \cellcolor[gray]{0.9}.529 & \cellcolor[gray]{0.9}.529 & .528 & .514 \\
\textbf{Sensationalism} & .563 & .571 & \cellcolor[gray]{0.9}.653 & \cellcolor[gray]{0.9}.653 & .581 & .581 & \cellcolor[gray]{0.9}.694 & \cellcolor[gray]{0.9}.645 \\
\textbf{News-worthiness} & \cellcolor[gray]{0.9}\underline{\textbf{.729}} & \cellcolor[gray]{0.9}\underline{\textbf{.711}} & \textbf{.683} & \textbf{.683} & \cellcolor[gray]{0.9}\textbf{.686} & \cellcolor[gray]{0.9}\textbf{.686} & .694 & .667 \\ \bottomrule[1pt]
\textbf{Overall} & .604 & .612 & \cellcolor[gray]{0.9}{.652} & \cellcolor[gray]{0.9}{.652} & .638 & .628 & \cellcolor[gray]{0.9}{\underline{\textbf{.705}}} & \cellcolor[gray]{0.9}{\underline{\textbf{.705}}} \\
\bottomrule[1pt]
\end{tabular}
\end{table}

\vspace{0.5em}
\noindent \textbf{E3:} \textit{Importance Analysis for Clickbait-related Features and Attributes.} Random forest is used to identify most important features, among which the top five features are presented in Table \ref{tab:clickbait_featureImportance}.
Results indicate that (1) headline readability, sensationalism and news-worthiness all play a role in differentiating fake news articles from the true ones; and (2) consistent with their performance in predicting fake news, features measuring news-worthiness of headlines rank relatively higher compared to that assessing headline readability and sensationalism.

\begin{table}[t]
\small
\caption{Important Clickbait-related Features and Attributes for Fake News Detection.}
\label{tab:clickbait_featureImportance}
\begin{threeparttable}
\begin{tabular}{@{}clc|lc@{}}
\toprule[1pt]
\multirow{2}{*}{\textbf{Rank}} & \multicolumn{2}{c|}{\textbf{PolitiFact}} & \multicolumn{2}{c}{\textbf{BuzzFeed}} \\ \cline{2-5} 
 & \textbf{Feature}  & \textbf{Attribute} & \textbf{Feature} & \textbf{Attribute}   \\ \hline
1 & \begin{tabular}[c]{@{}l@{}}Similarity (\textsc{\textsc{word2vec}})\end{tabular}       & S/N   
  & \begin{tabular}[c]{@{}l@{}}Similarity (\textsc{word2vec})\end{tabular}   & S/N       \\
2 & \begin{tabular}[c]{@{}l@{}}Similarity (\textsc{Sentence2Vec})\end{tabular}       & S/N     
  & \# Characters                                                       & R       \\
3 & \% Netspeak                                                         & N   
  & \# Words                                                            & R         \\
4 & Sentiment Polarity                                                  & S     
  & \# Syllables                                                        & R         \\
5 & Coleman-Liau Index                                                                 & R     
  & Gunning-Fog Index                                                                  & R         \\   
  \bottomrule[1pt]
\end{tabular}
\begin{tablenotes}
\item R: Readability; S: Sensationalism; N: News-worthiness
\end{tablenotes}
\end{threeparttable}
\end{table}

\begin{figure}[t]
    \centering
    \subfigure[PolitiFact]{
    \includegraphics[width=0.3\textwidth]{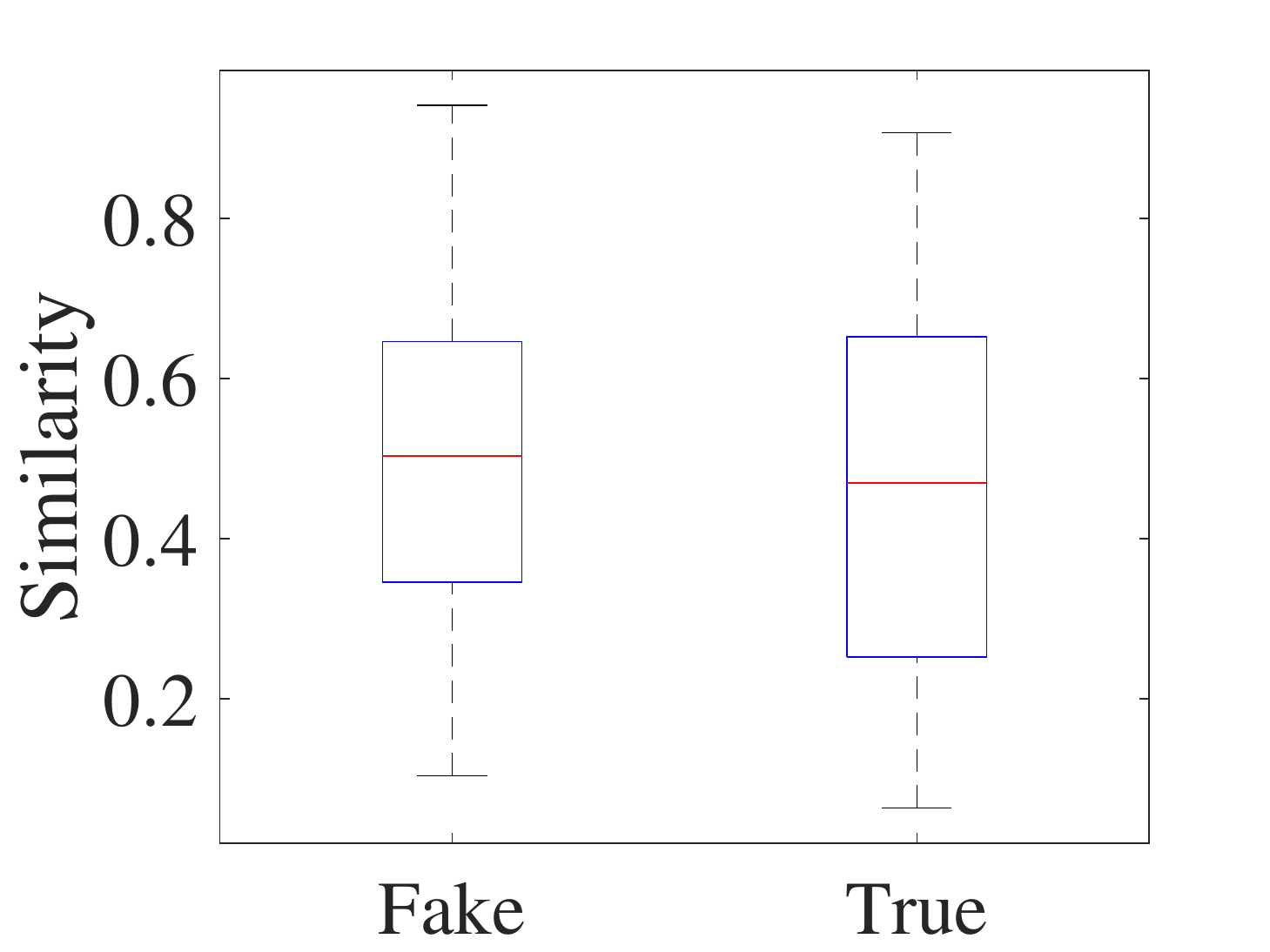}
    \includegraphics[width=0.3\textwidth]{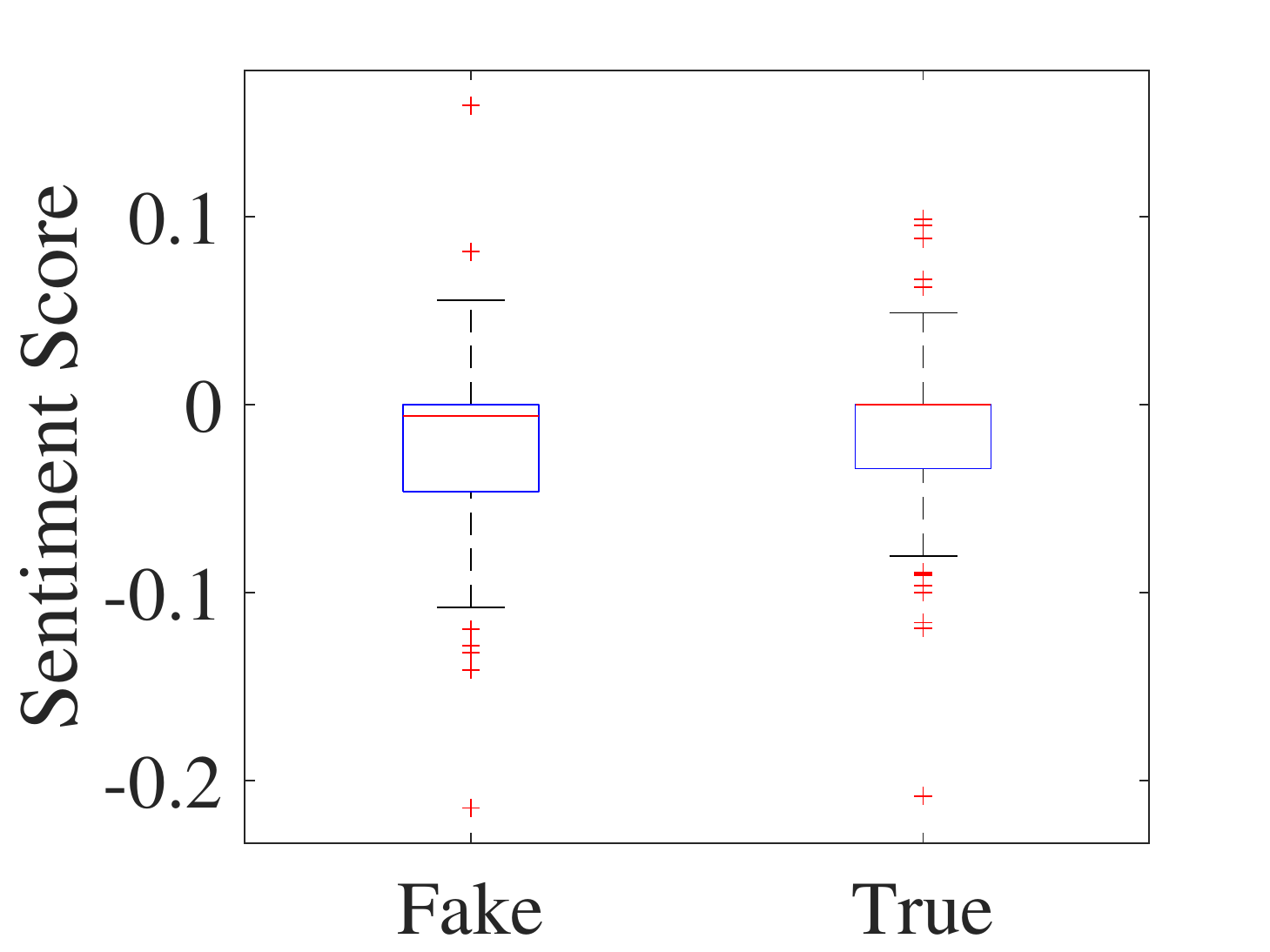}
    \includegraphics[width=0.3\textwidth]{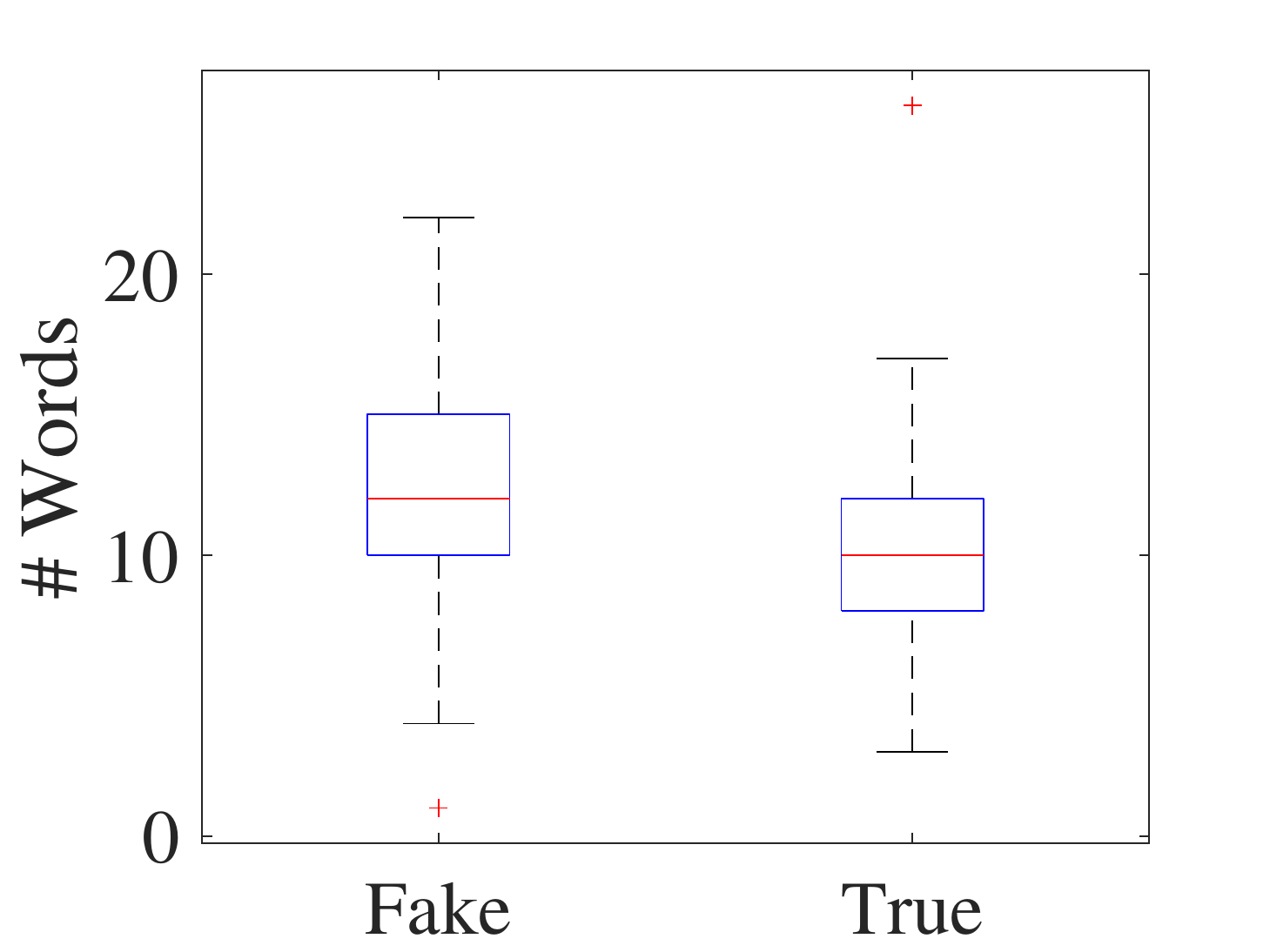}} 
    \subfigure[BuzzFeed]{
    \includegraphics[width=0.3\textwidth]{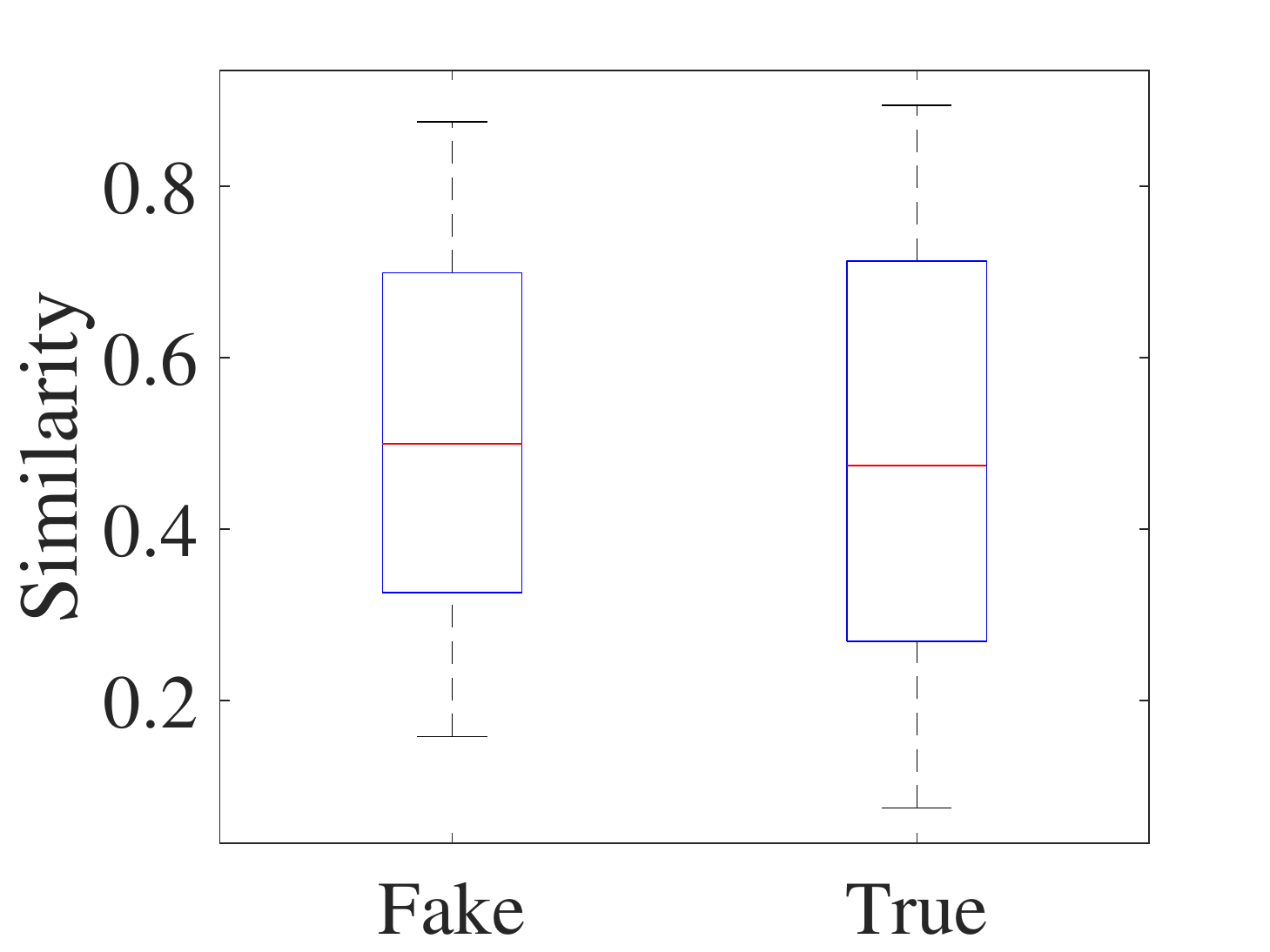}
    \includegraphics[width=0.3\textwidth]{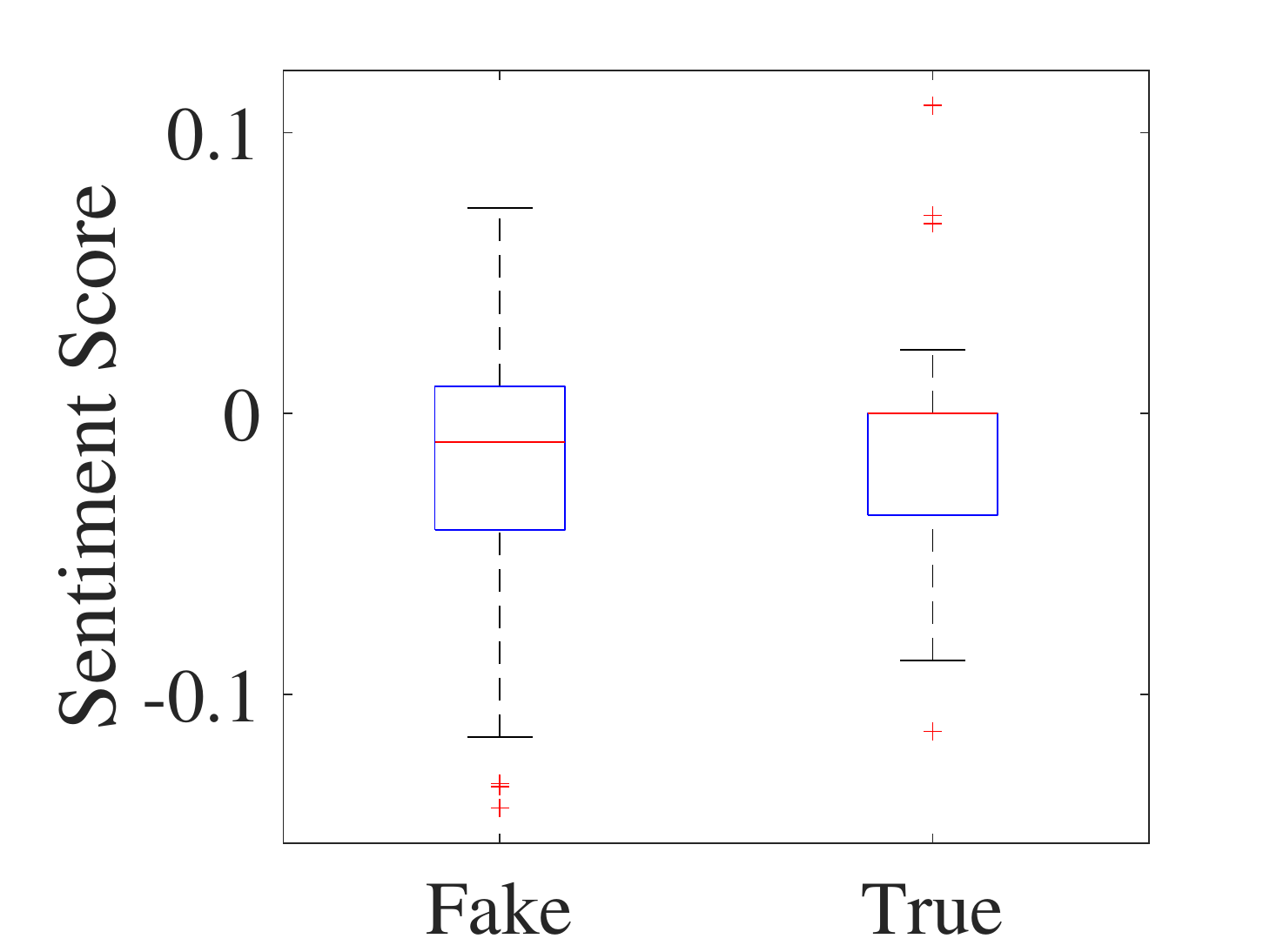}
    \includegraphics[width=0.3\textwidth]{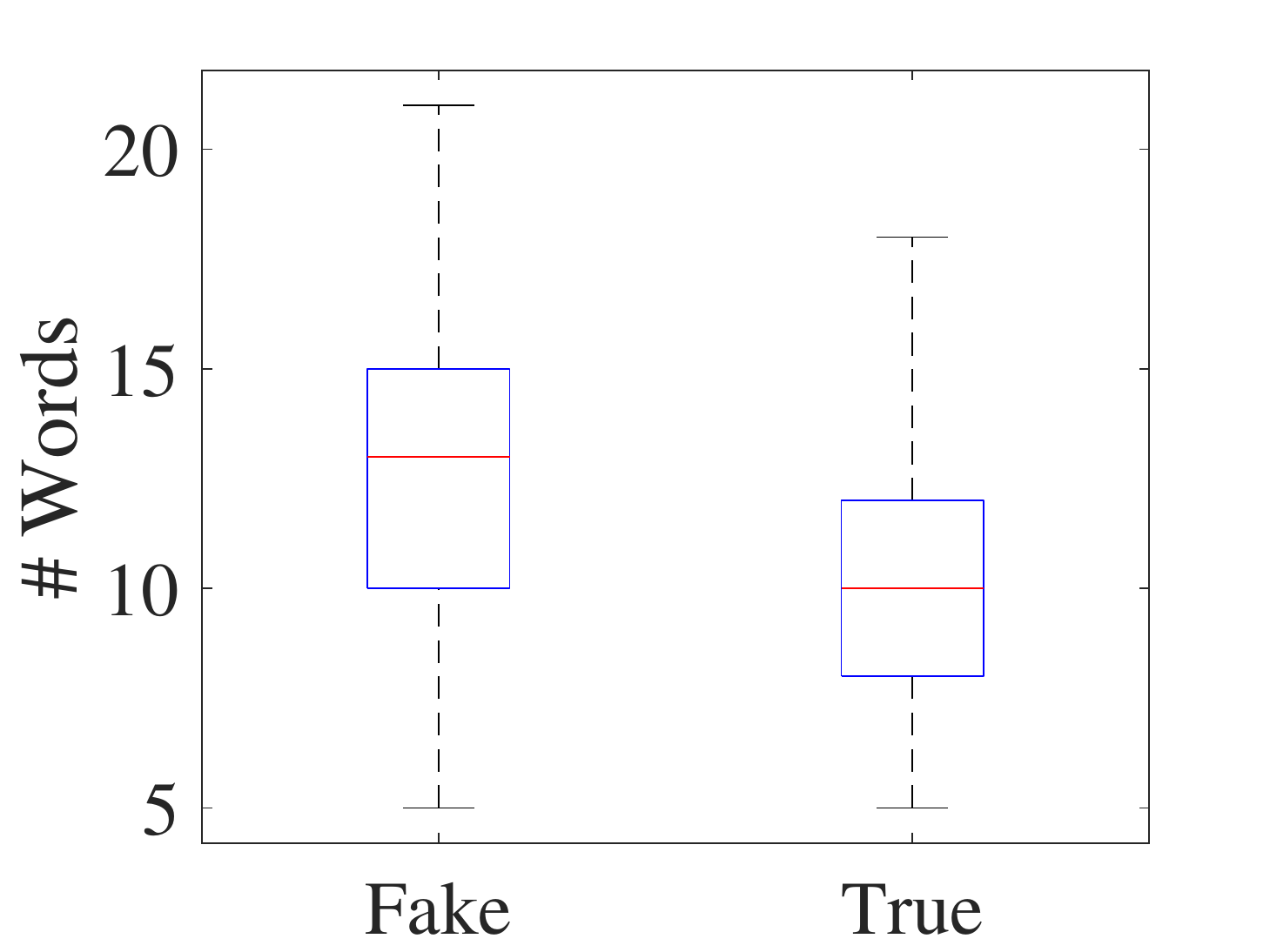}}
    \caption{Potential Patterns of Fake News Headlines. In both datasets, fake news headlines are generally (i) less similar to their body-text, and contain (ii) more words when compared to true news. In addition, (iii) fake news headlines are slightly inclined to be negative with a broader scope of sentiment scores; while true news are comparatively neutral with a more narrow scope of sentiment scores.}
    \label{fig:clickbait_pattern}
    \end{figure}

\vspace{0.5em}
\noindent \textbf{E4:} \textit{Potential Patterns of Fake News Headlines.} 
Using the boxplot of clickbait features within fake and true news, we examine whether fake news headlines share some potential patterns with clickbaits. Results are provided in Figure \ref{fig:clickbait_pattern}.
Specifically,

\begin{itemize}
    \item Figures on the left column present the box-plot of the cosine similarity between news headlines and their corresponding body-text, which is computed using the \textsc{Sentence2Vec} model~\cite{arora2016simple}.
    Such similarity is assumed to be positively correlated to the sensationalism and negatively correlated to the news-worthiness of news headlines. Both figures reveal that, in general, fake news headlines are less similar to their body-text compared to true news headlines, which matches with the characteristic of clickbaits~\cite{dong2019similarity}.
    
    \item Figures on the middle column present the box-plot of the average sentiment score of words within a news headline. Both figures reveal that,
    in general, fake news headlines are slightly inclined to be negative with a broader scope of sentiment scores; while true news are comparatively neutral with a more narrow scope of sentiment scores. In other words, fake news headlines are more likely to be negative, or to be sensational with an extreme emotion, which matches with the characteristic of clickbaits~\cite{chakraborty2016stop}.
    
    \item Figures on the right column present the box-plot of the number of words within news headlines, as one of the parameters of readability criteria and features representing news readability. Though it cannot directly measure the readability of news headlines, we find that fake news headlines often contain more words (as well as syllables and characters) compared to true news. 
\end{itemize}

\begin{figure}[t]
\centering
\subfigure[PolitiFact]{ \label{subfig:dataDistribution_pf}
\includegraphics[width=.46\textwidth]{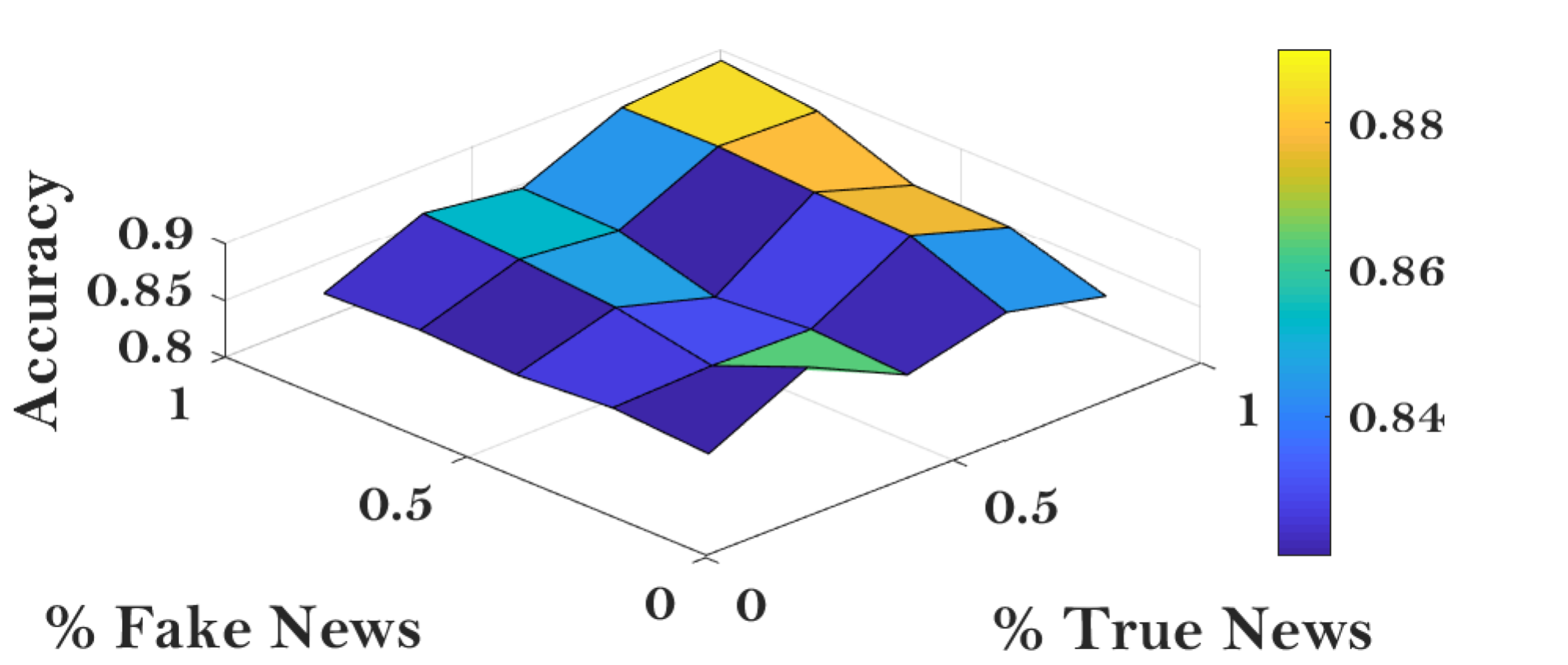} 
\includegraphics[width=.46\textwidth]{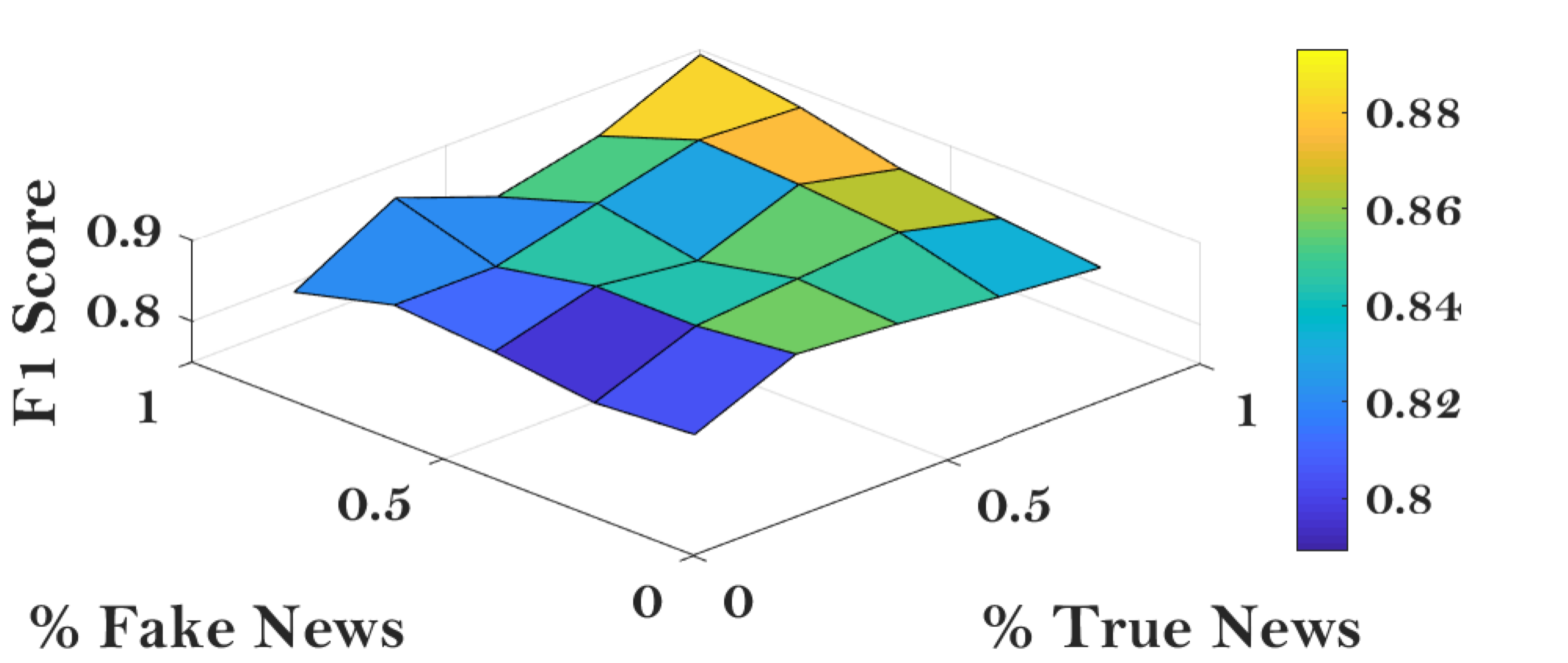}} 
\subfigure[BuzzFeed]{ \label{subfig:dataDistribution_bf}
\includegraphics[width=.46\textwidth]{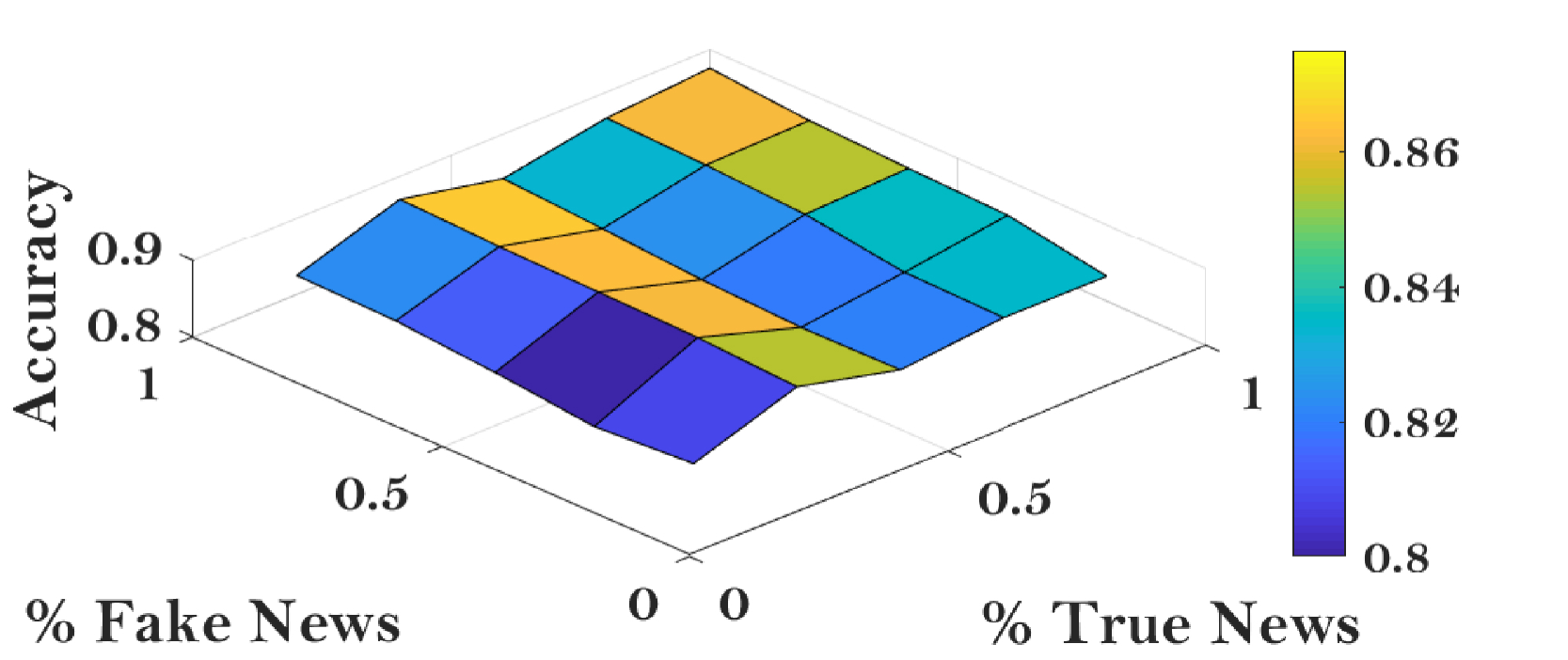} 
\includegraphics[width=.46\textwidth]{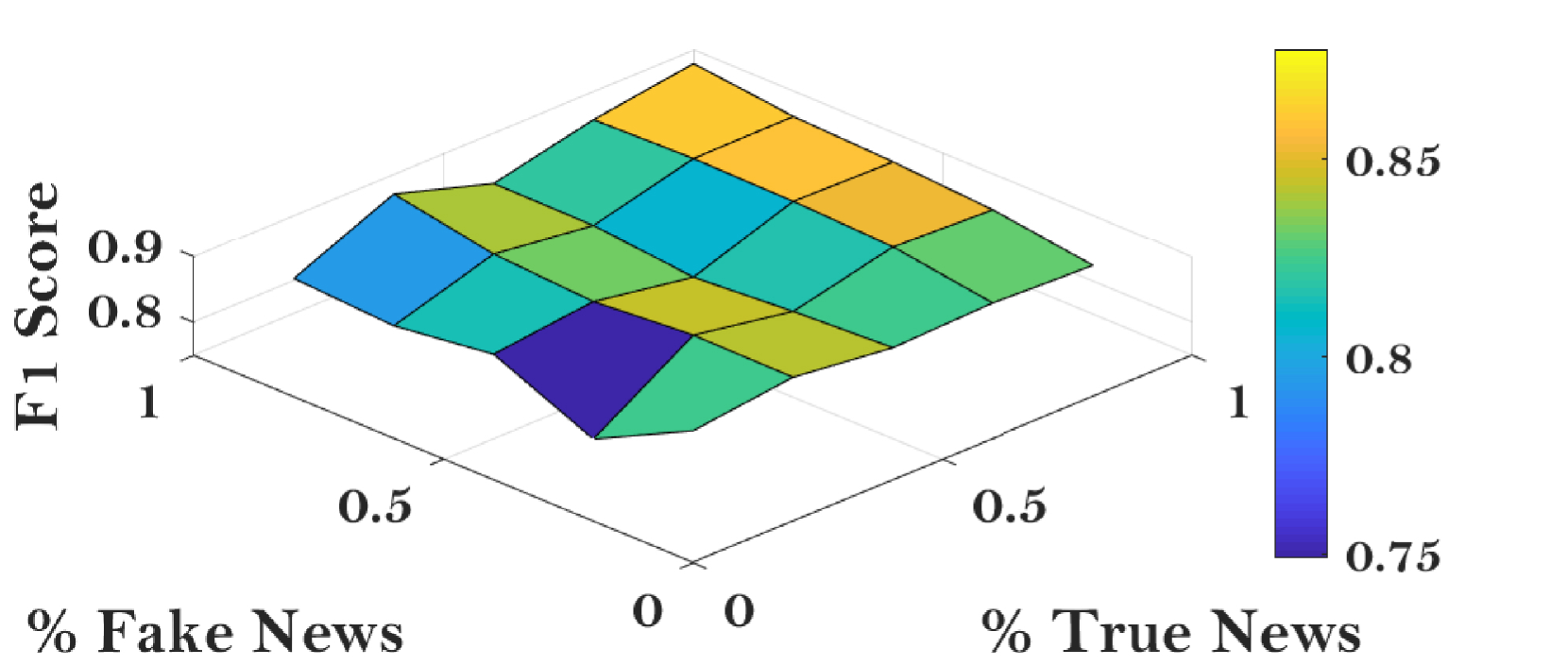}}
\caption{Performance Sensitivity to News Distribution (\% Fake News vs. \% True News)}
\label{fig:dataDistribution}

\centering
\subfigure[PolitiFact]{ \label{subfig:trainingNewsProp_pf}
\includegraphics[width=.33\textwidth]{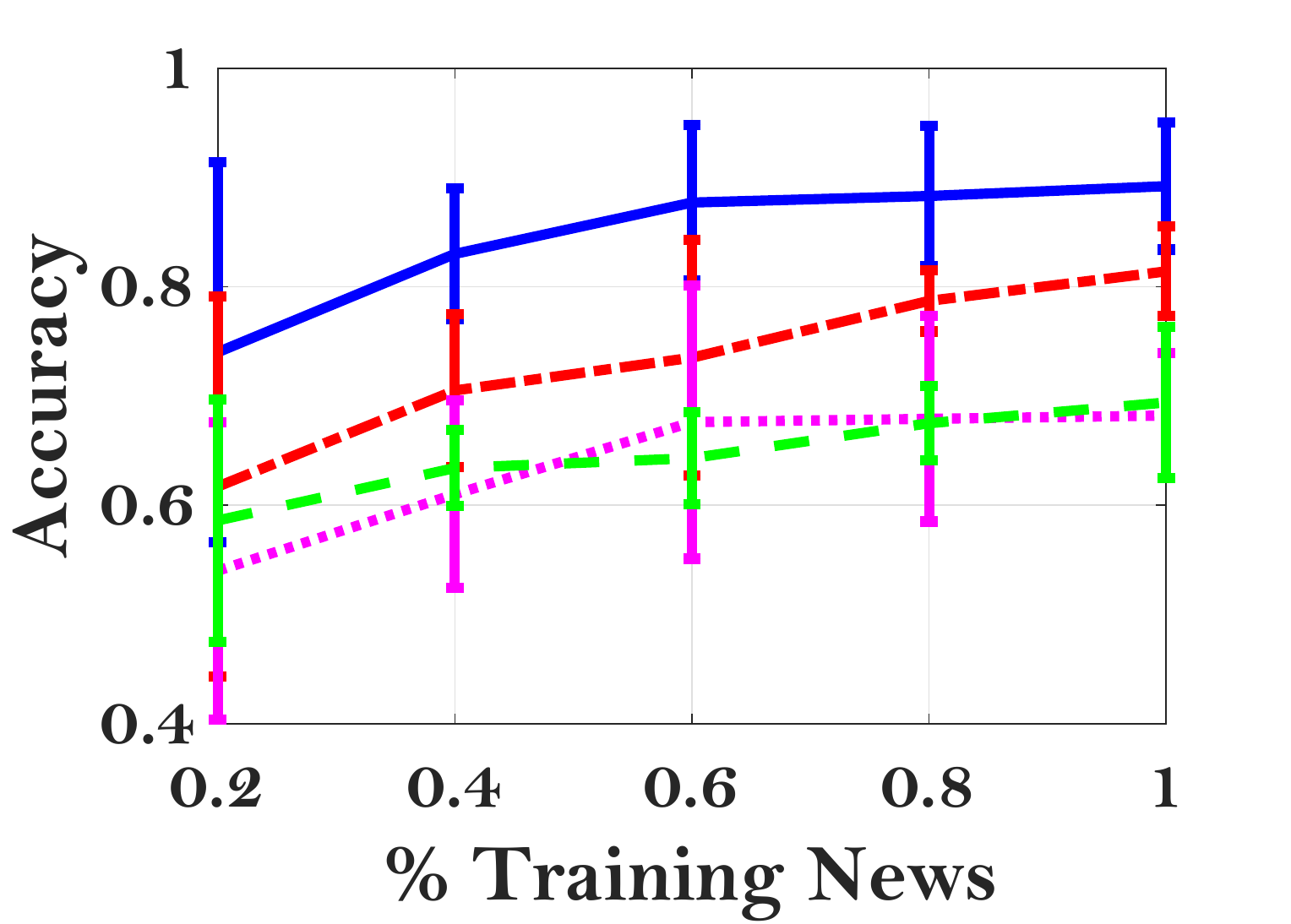} \quad
\includegraphics[width=.51\textwidth]{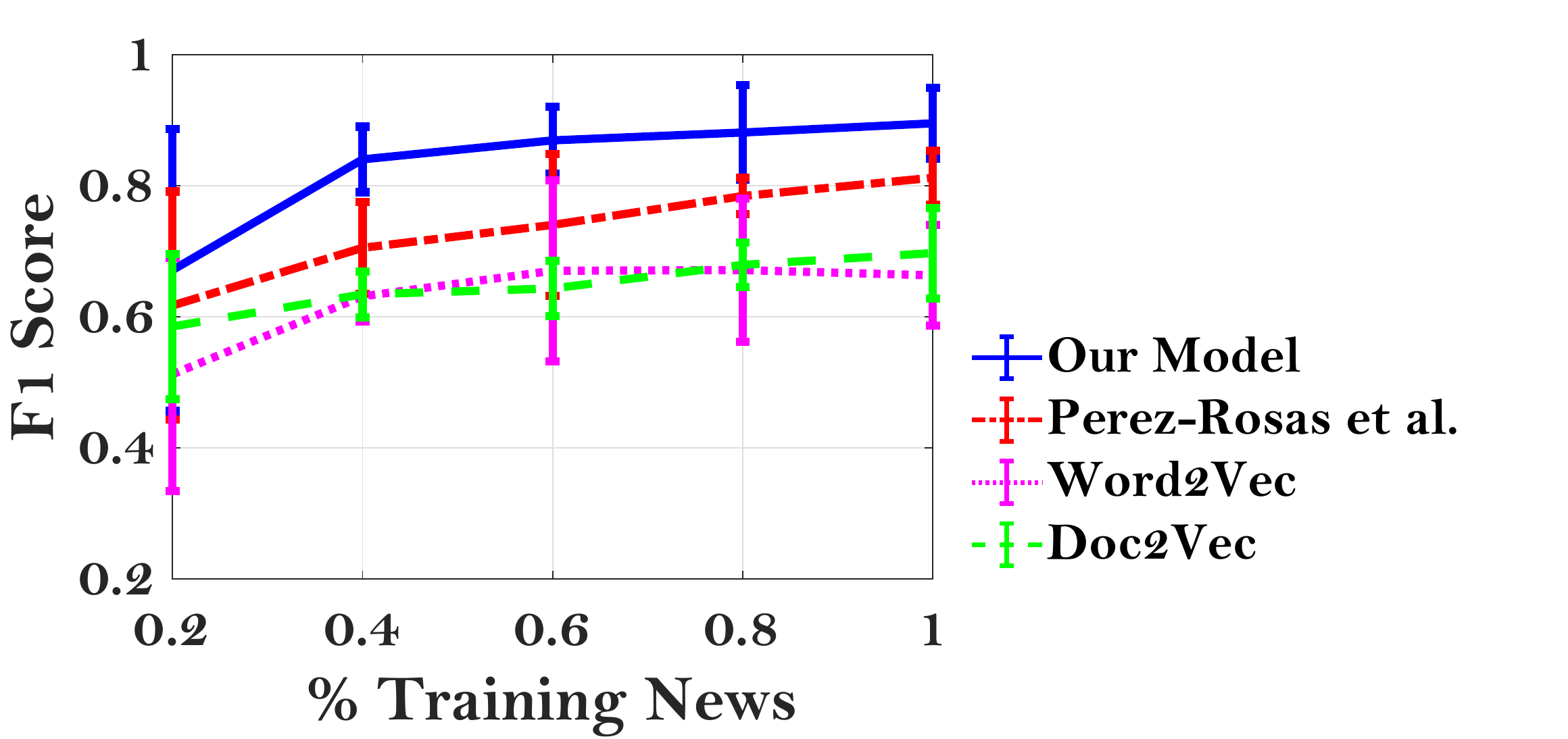}} 
\subfigure[BuzzFeed]{ \label{subfig:trainingNewsProp_bf}
\includegraphics[width=.33\textwidth]{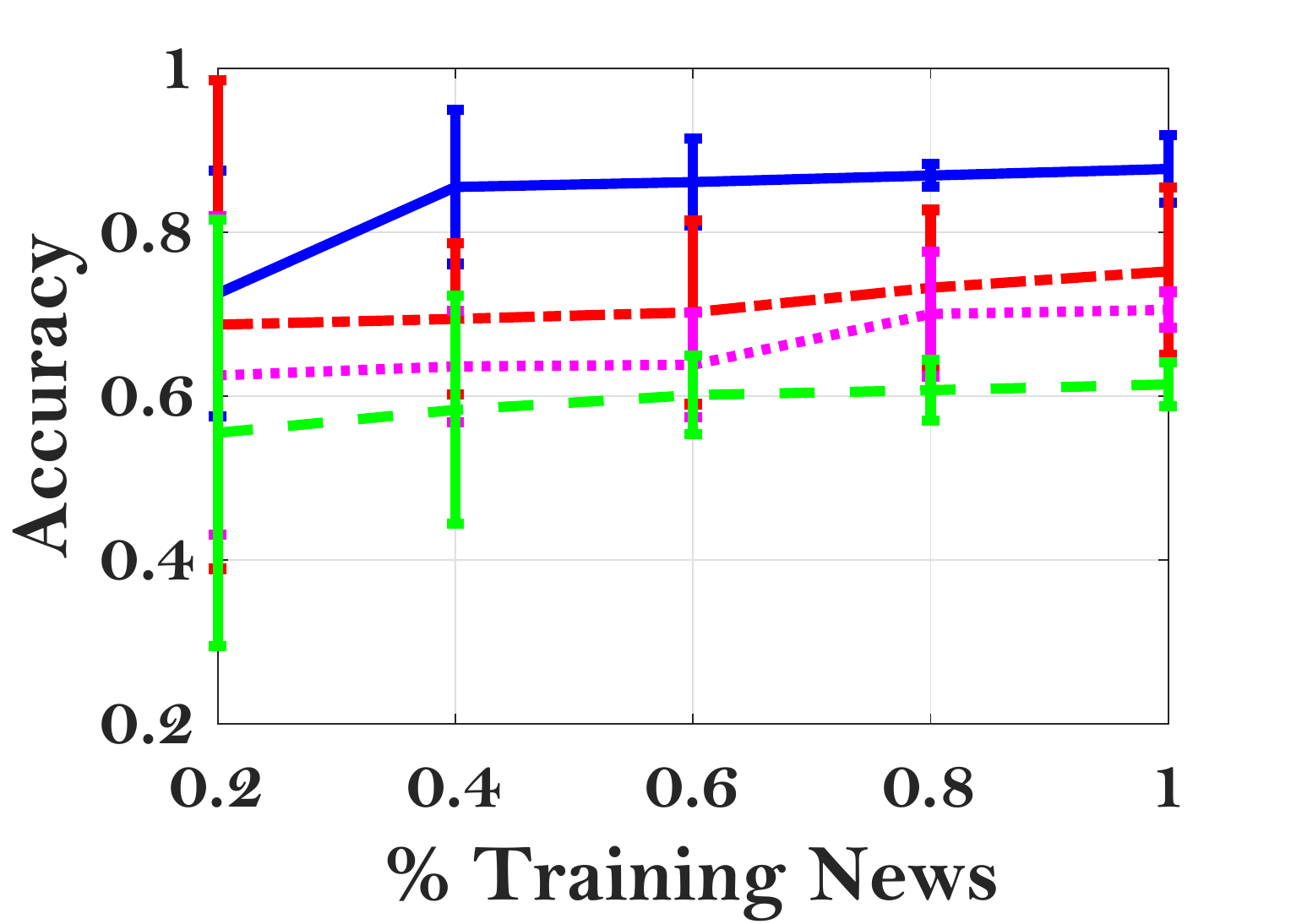} \quad
\includegraphics[width=.51\textwidth]{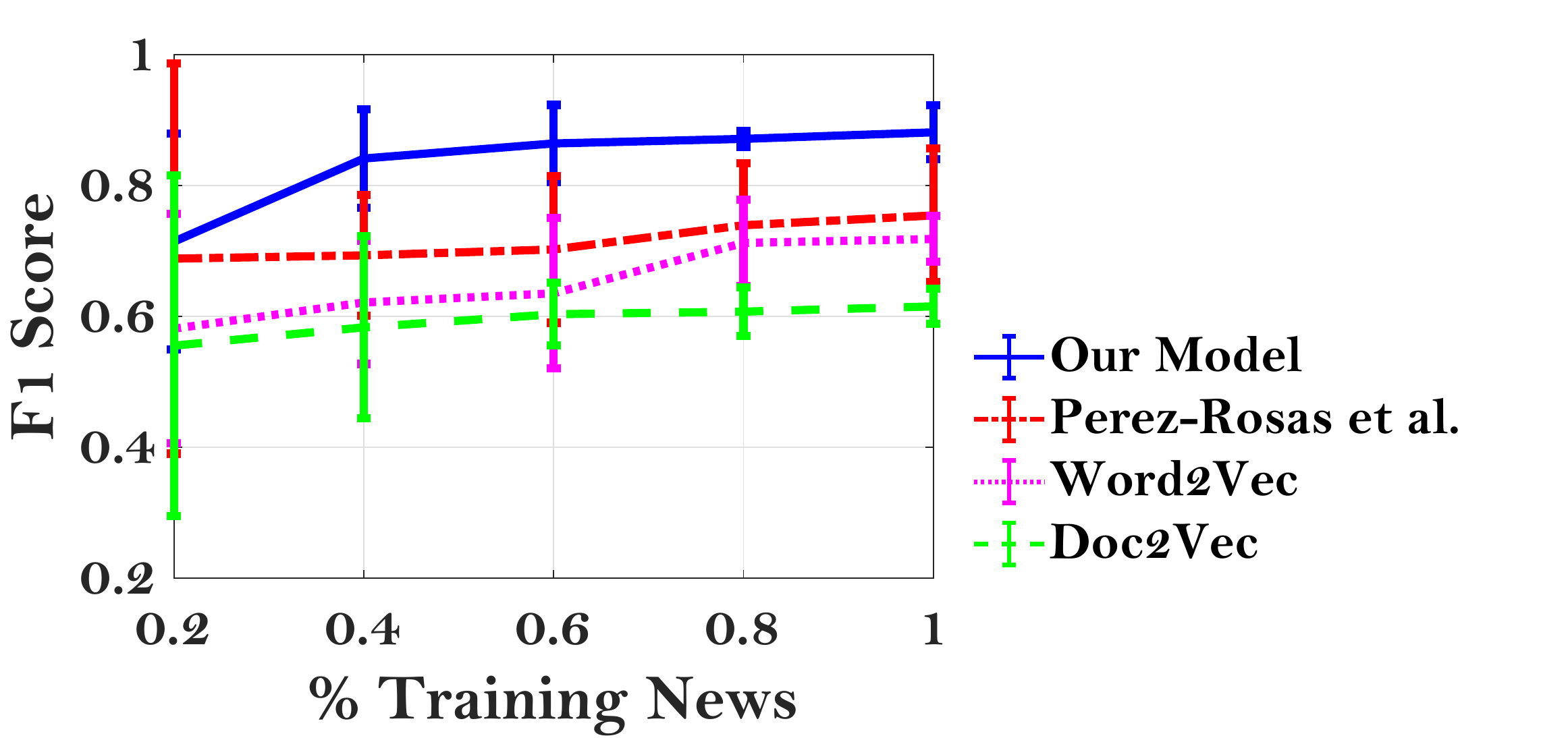}}
\caption{Impact of the Number of Training News Articles in Predicting Fake News.}
\label{fig:trainingNewsProp}
\end{figure}

\subsubsection{Impact of News Distribution on Fake News Detection}
\label{subsubsec:newsDistribution}
We assess the sensitivity of our model to the news distribution, i.e., the proportion of true vs. fake news stories within the population, which are initially equal in both PolitiFact and BuzzFeed datasets. Specifically, we randomly select a proportion ($\in (0,1]$) of fake news stories and a proportion of true news stories in each dataset. The corresponding accuracy and $F_1$ scores by using XGBoost are presented in Figure \ref{fig:dataDistribution}.
Results on both datasets indicate that the performance of the proposed model fluctuates between $\sim$0.75 and $\sim$0.9. However, in most cases, the model is resilient to such perturbations and the accuracy and $F_1$ scores are between $\sim$0.8 and $\sim$0.88.

\subsubsection{Fake News Early Detection} Compared to propagation-based models, content-based fake news detection models can detect fake news before it has been disseminated on social media. Among content-based fake news detection models, their early detection ability also depends on how much prior knowledge they require to accurately detect fake news~\cite{zhou2018survey,wang2018eann}. Here, we measure the amount of such prior knowledge from two perspectives: (\textbf{E1}) the number of news articles available for learning and training a classifier, and (\textbf{E2}) the content quantity for each news article available for training and predicting fake news.
\label{subsubsec:early}

\vspace{0.5em}
\noindent \textbf{E1:} \textit{Model Performance with Limited Number of Training News Articles.} In this experiment, we randomly select a proportion ($\in (0,1]$) of news articles from each of the PolitiFact and BuzzFeed datasets. Performance of several content-based models in predicting fake news is then evaluated based on the selected subset of news articles, which has been presented in Figure \ref{fig:trainingNewsProp}. It can be observed from Figure \ref{fig:trainingNewsProp} that with the change of the number of available training news articles, the proposed model performs best in most cases. 
Note that, compared to random sampling, sampling based on the time that news articles were published is a more proper strategy when evaluating the early detection ability of models; however, such temporal information has not been fully provided in the datasets.

\begin{figure}[t]
\centering
\subfigure[PolitiFact]{ \label{subfig:newsContentSize_pf}
\includegraphics[width=.34\textwidth]{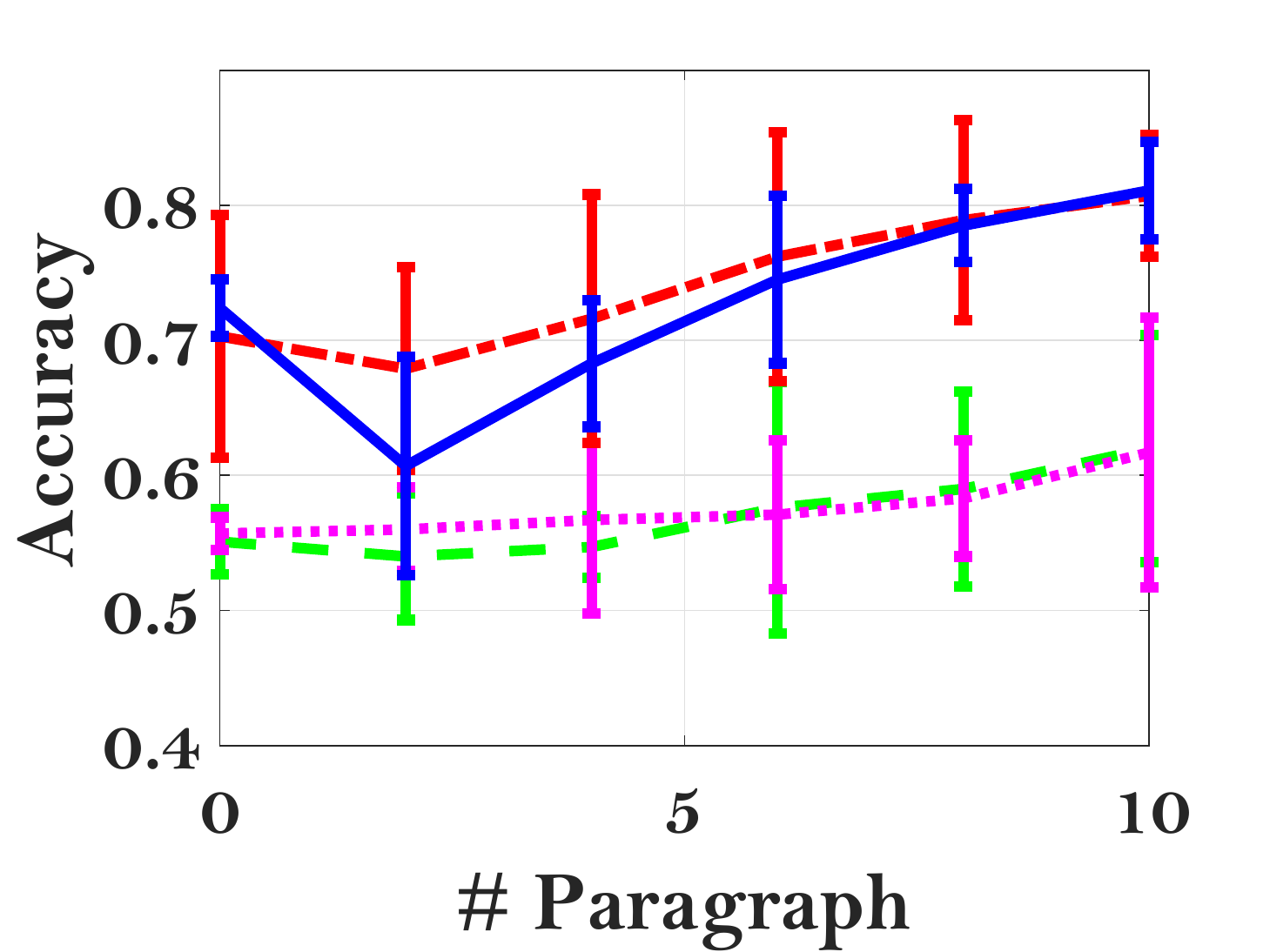} \quad
\includegraphics[width=.52\textwidth]{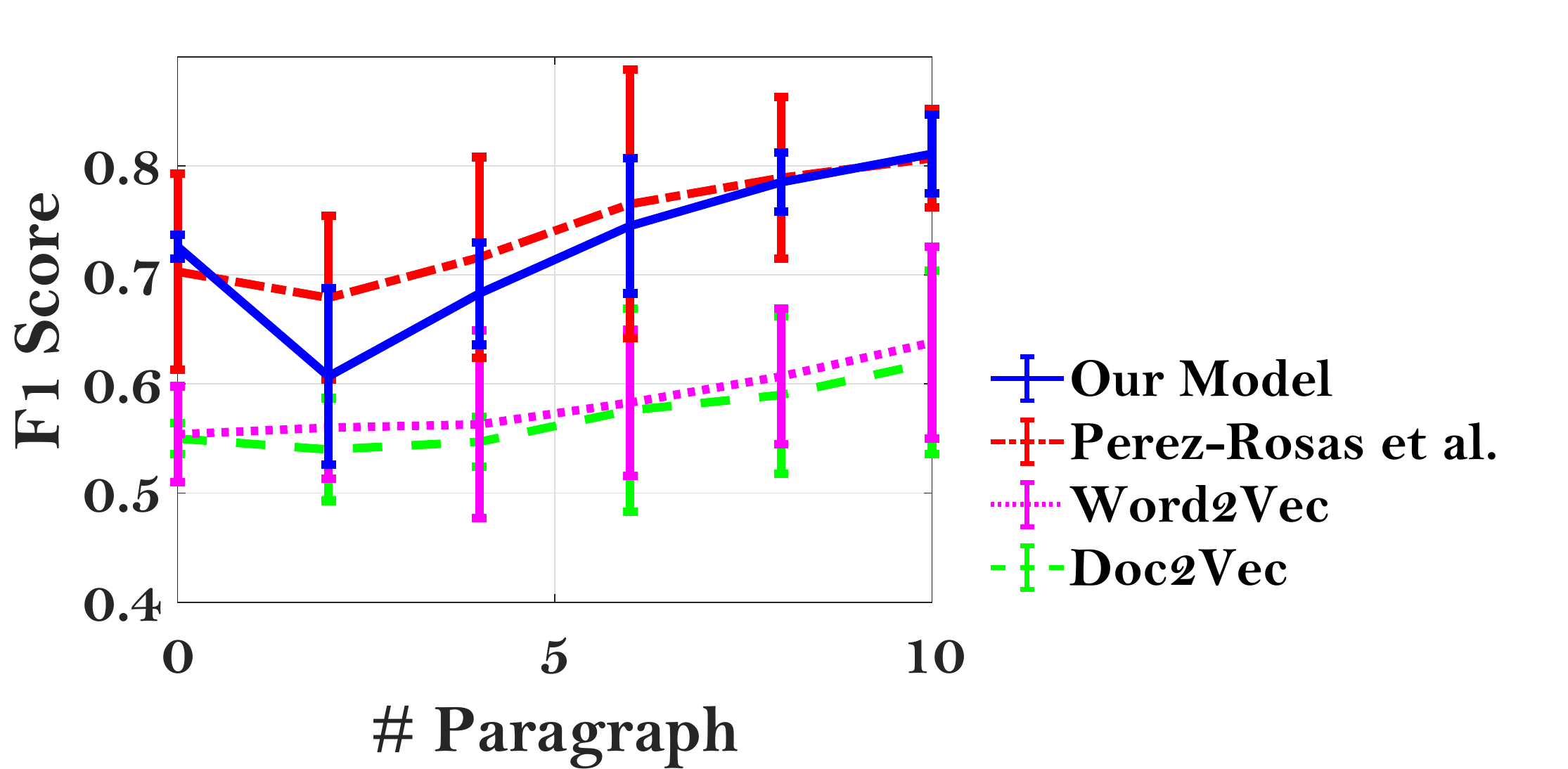}} 
\subfigure[BuzzFeed]{ \label{subfig:newsContentSize_bf}
\includegraphics[width=.34\textwidth]{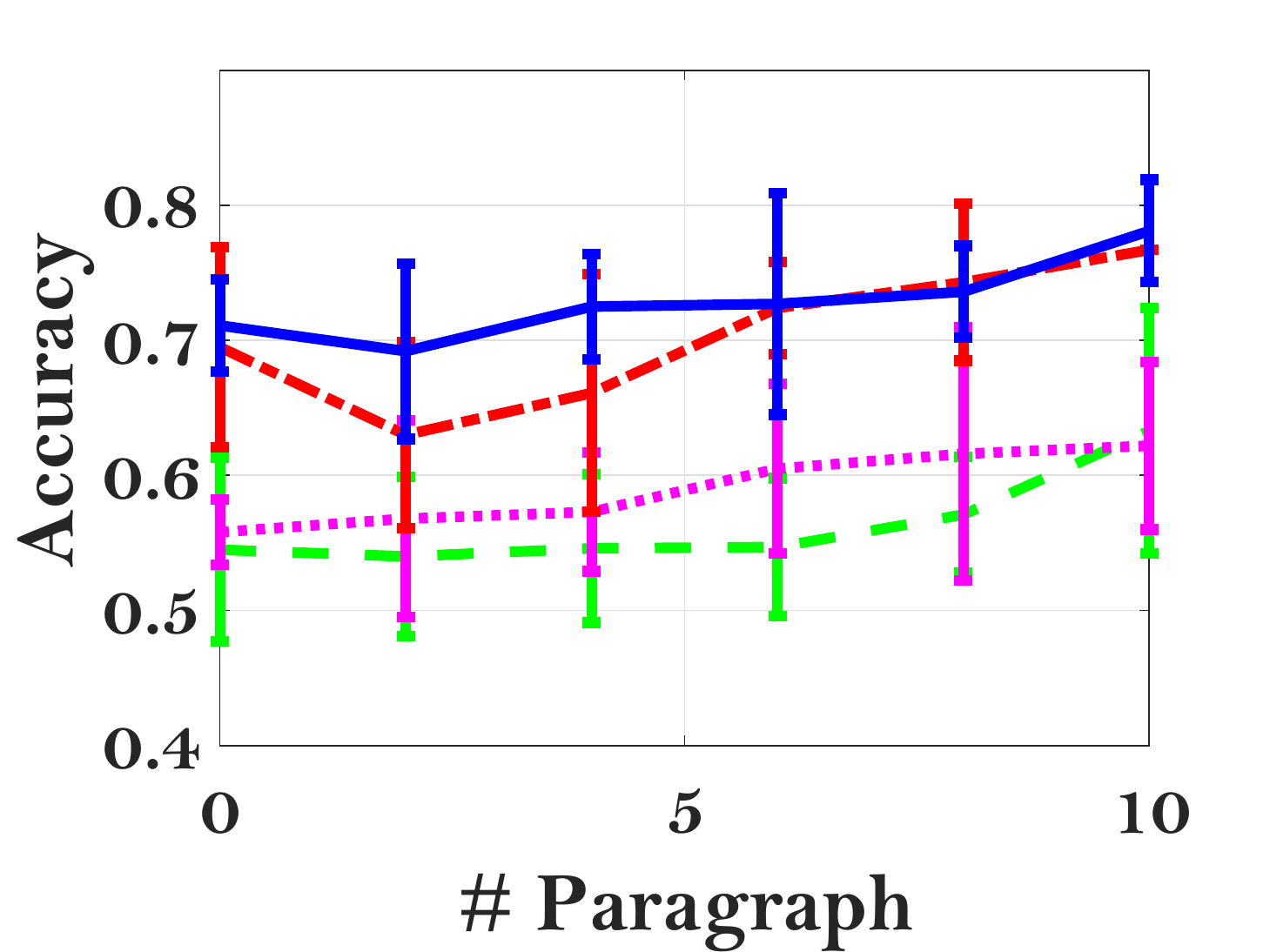} \quad
\includegraphics[width=.52\textwidth]{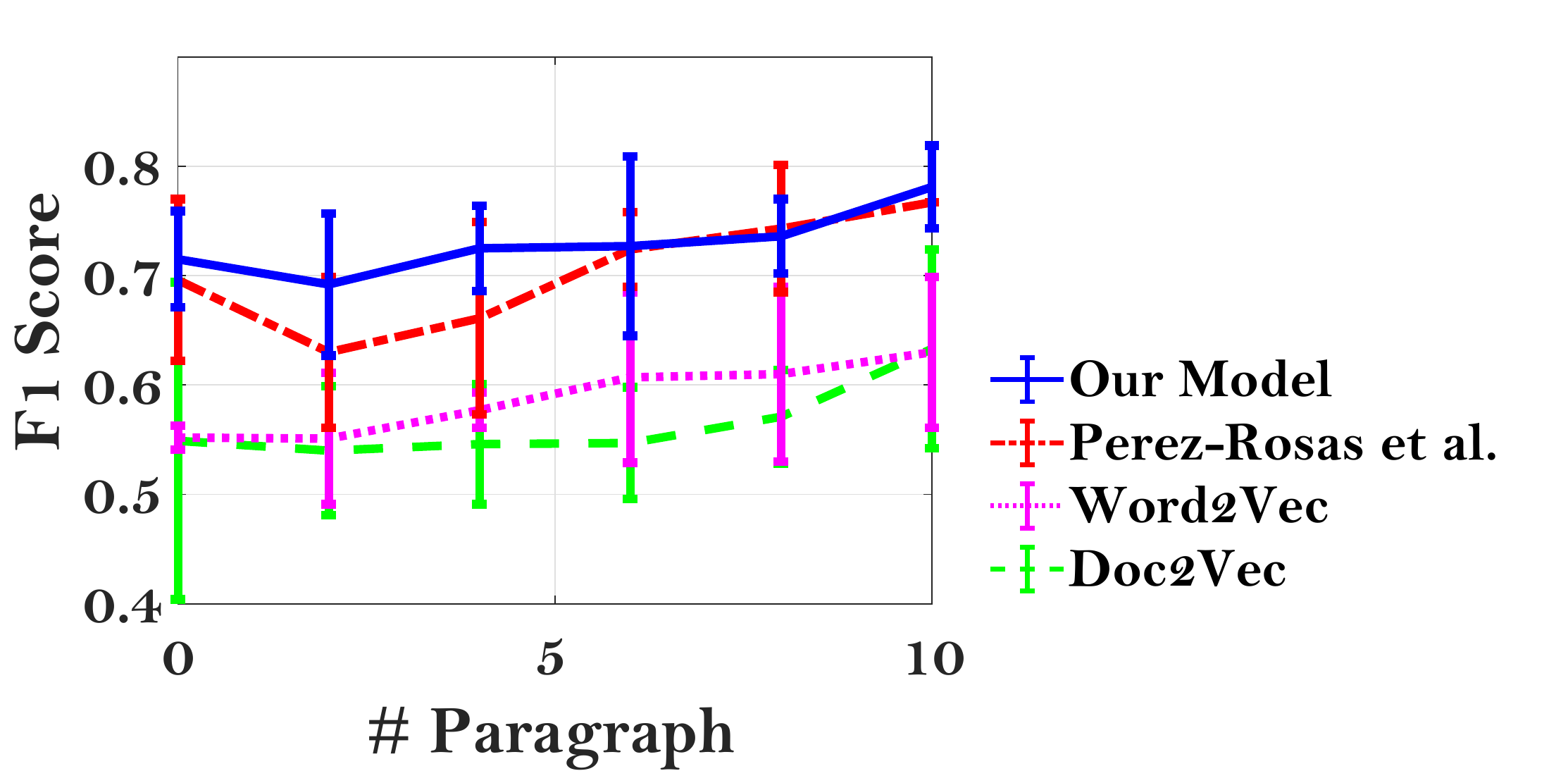}}
\caption{Impact of the Available Information within News Content in Predicting Fake News.}
\label{fig:newsContentSize}
\end{figure}

\vspace{0.5em}
\noindent \textbf{E2:} \textit{Model Performance with Limited News Content Information.} In this experiment, we assess the performance of our fake news model when partial news content information is available. Specifically, such partial news content information ranges from the headline of the news article to the headline with $n$ ($n=1,2,\cdots$) randomly selected paragraph(s) from the article. Results are presented in Figure \ref{fig:newsContentSize}, which indicate that (1) compared to the linguistic model proposed by Perez-Rosas et al.~\cite{perez2017automatic}, our model generally has a comparable performance while can always outperform it when only news headline information is available (i.e., \# paragraphs is 0); and (2) our model can always perform better than the models based on the latent representation of news content~\cite{le2014distributed,mikolov2013efficient}.

\section{Conclusion}
\label{sec:conclusion}

In this paper, an interdisciplinary study is conducted  for explainable fake news early detection. To predict fake news before it starts to propagate on social media, our work comprehensively studies and represents news content at four language levels: lexicon-level, syntax-level, semantic-level, and discourse-level. Such representation is inspired by well-established theories in social and forensic psychology. Experimental results based on real-world datasets indicate that the performance (i.e., accuracy and $F_1$ score) of the proposed model can (1) generally achieve  $\sim$88\%, outperforming all baselines which include content-based, propagation-based and hybrid (content+propagation) fake news detection models; and (2) maintain $\sim$80\% and $\sim$88\% when data size and news distribution (\% fake news vs. \% true news) vary. Among content-based models, we observe that (3) the proposed model performs comparatively well in predicting fake news with limited prior knowledge. We also observe that (4) similar to deception, fake news differs in content style, quality and sentiment from the truth, while carries similar levels of cognitive and perceptual information compared to the truth. (5) Similar to clickbaits, fake news headlines present higher sensationalism and lower news-worthiness while their readability characteristics are complex and difficult to be directly concluded. In addition, fake news (6) is often matched with shorter words and longer sentences.
With three stages of being created, being published on news outlet(s), and being propagated on any social media (medium), based on the proposed method, fake news proliferation can be mitigated before social media users have touched with it. Meanwhile, we should emphasize that a news article is likely, based on our observations, to be fake when it matches all (instead of any) potential patterns in its content. Note that our results do not indicate any news article sharing such characteristics is absolutely fake. 
To systematically reveal further patterns in fake news content compared to true news content one has to involve (1) more fundamental theories and (2) empirical analyses on larger real-world datasets (see \cite{vosoughi2018spread} for an illustrated analysis for fake news propagation and see \cite{norregaard2019nela} for a recently released large-scale dataset). Datasets consisting of the ground truth of, e.g., both fake news and clickbaits, are invaluable to understand the relationships among different types of unreliable information; while such datasets are so far rarely available. Furthermore, it should be pointed out that effective utilization of rhetorical relationships and utilizing news images~\cite{parikh2018media} in an explainable way for fake news detection are still open issues. All aforementioned limitations will be part of our future work.


\bibliographystyle{ACM-Reference-Format}
\bibliography{reference2}

\begin{appendix}
\section{Semantic-level Features}
\label{app:semantic_features}

Table \ref{tab:semantic_features} provides a detailed list of semantic-level features involved in our study.

\begin{small}
\begin{longtable}{|c|c|c|l|c|}
\caption{Semantic-level Features}
\label{tab:semantic_features} \\
\hline & \multicolumn{2}{c|}{\textbf{Attribute}} & \textbf{Feature(s)} & \textbf{Tool \& Ref.} \\
\hline
\multirow{40}{*}{\rotatebox{90}{\textbf{Disinformation-related Attributes (DIAs) (72)}}
}& \multirow{15}{*}{\rotatebox{90}{Quality (30)}} & \multirow{6}{*}{Informality (12)} & \#/\% Swear Words & \multirow{6}{*}{LIWC} \\ \cline{4-4}
 &  &  & \#/\% Netspeak &  \\ \cline{4-4}
 &  &  & \#/\% Assent &  \\ \cline{4-4}
 &  &  & \#/\% Nonfluencies &  \\ \cline{4-4}
 &  &  & \#/\% Fillers &  \\ \cline{4-4}
 &  &  & Overall \#/\% Informal Words &  \\ \cline{3-5} 
 &  & \multirow{6}{*}{Diversity (12)} & \#/\% Unique Words & Self-implemented \\ \cline{4-5} 
 &  &  & \#/\% Unique Content Words & LIWC \\ \cline{4-5} 
 &  &  & \#/\% Unique Nouns & \multirow{4}{*}{\begin{tabular}[c]{@{}c@{}}NLTK\\ POS Tagger\end{tabular}} \\ \cline{4-4}
 &  &  & \#/\% Unique Verbs &  \\ \cline{4-4}
 &  &  & \#/\% Unique Adjectives &  \\ \cline{4-4}
 &  &  & \#/\% Unique Adverbs &  \\ \cline{3-5} 
 &  & \multirow{3}{*}{Subjectivity (6)} & \#/\% Biased Lexicons & \multirow{2}{*}{\cite{recasens2013linguistic}} \\ \cline{4-4}
 &  &  & \#/\% Report Verbs &  \\ \cline{4-5} 
 &  &  & \#/\% Factive Verbs & \cite{hooper1975assertive} \\ \cline{2-5} 
 & \multicolumn{2}{c|}{\multirow{7}{*}{Sentiment (13)}} & \#/\% Positive Words & \multirow{6}{*}{LIWC} \\ \cline{4-4}
 & \multicolumn{2}{c|}{} & \#/\% Negative Words &  \\ \cline{4-4}
 & \multicolumn{2}{c|}{} & \#/\% Anxiety Words &  \\ \cline{4-4}
 & \multicolumn{2}{c|}{} & \#/\% Anger Words &  \\ \cline{4-4}
 & \multicolumn{2}{c|}{} & \#/\% Sadness Words &  \\ \cline{4-4}
 & \multicolumn{2}{c|}{} & Overall \#/\% Emotional Words &  \\ \cline{4-5} 
 & \multicolumn{2}{c|}{} & Avg. Sentiment Score of Words & \begin{tabular}[c]{@{}c@{}}NLTK.Sentiment\\ Package\end{tabular} \\ \cline{2-5} 
 & \multicolumn{2}{c|}{\multirow{7}{*}{Quantity (7)}} & \# Characters & Self-implemented \\ \cline{4-5} 
 & \multicolumn{2}{c|}{} & \# Words & Self-implemented \\ \cline{4-5} 
 & \multicolumn{2}{c|}{} & \# Sentences & Self-implemented \\ \cline{4-5} 
 & \multicolumn{2}{c|}{} & \# Paragraphs & Self-implemented \\ \cline{4-5} 
 & \multicolumn{2}{c|}{} & Avg. \# Characters Per Word & Self-implemented \\ \cline{4-5} 
 & \multicolumn{2}{c|}{} & Avg. \# Words Per Sentence & Self-implemented \\ \cline{4-5} 
 & \multicolumn{2}{c|}{} & Avg. \# Sentences Per Paragraph & Self-implemented \\ \cline{2-5} 
 &  &  & \#/\% Insight &  \\ \cline{4-4}
 &  &  & \#/\% Causation &  \\ \cline{4-4}
 &  &  & \#/\% Discrepancy &  \\ \cline{4-4}
 &  &  & \#/\% Tentative &  \\ \cline{4-4}
 & \multirow{5}{*}{\rotatebox{90}{Specificity (22)}} & \multirow{1}{*}{\begin{tabular}[c]{@{}c@{}}Cognitive Process (14)\end{tabular}} & \#/\% Certainty & \multirow{1}{*}{LIWC}  \\ \cline{4-4}
 &  &  & \#/\% Differentiation &  \\ \cline{4-4}
 &  &  & Overall \#/\% Cognitive Processes &  \\ \cline{3-4}
 &  & \multirow{4}{*}{\begin{tabular}[c]{@{}c@{}}Perceptual Process (8)\end{tabular}} & \#/\% See &  \\ \cline{4-4}
 &  &  & \#/\% Hear &  \\ \cline{4-4}
 &  &  & \#/\% Feel &  \\ \cline{4-4}
 &  &  & Overall \#/\% Perceptual Processes &  \\ \hline
 & \multicolumn{2}{c|}{\multirow{3}{*}{\begin{tabular}[c]{@{}c@{}}General Clickbait\\Patterns (3)\end{tabular}}} & \# Common Clickbait Phrases & \multirow{3}{*}{\cite{gianotto2014downworthy}} \\ \cline{4-4}
 & \multicolumn{2}{c|}{} & \# Common Clickbait Expressions &  \\ \cline{4-4}
 & \multicolumn{2}{c|}{} & Overall \# Common Clickbait Patterns &  \\ \cline{2-5} 
 & \multicolumn{2}{c|}{} & Flesch Reading Ease Index (FREI) & Self-implemented \\ \cline{4-5} 
 & \multicolumn{2}{c|}{} & Flesch-Kioncaid Grade Level (FKGL) & Self-implemented \\ \cline{4-5} 
 & \multicolumn{2}{c|}{} & Automated Readability Index (ARI) & Self-implemented \\ \cline{4-5} 
 & \multicolumn{2}{c|}{} & Gunning Fox Index (GFI) & Self-implemented \\ \cline{4-5} 
 & \multicolumn{2}{c|}{} & Coleman-Liau Index (CLI) & Self-implemented \\ \cline{4-5} 
 & \multicolumn{2}{c|}{} & \# Words & Self-implemented \\ \cline{4-5} 
 & \multicolumn{2}{c|}{} & \# Syllables & Self-implemented \\ \cline{4-5} 
 & \multicolumn{2}{c|}{} & \# Polysyllables & Self-implemented \\ \cline{4-5} 
 & \multicolumn{2}{c|}{} & \# Characters & Self-implemented \\ \cline{4-5} 
 & \multicolumn{2}{c|}{\multirow{-10}{*}{Readability (10)}} & \# Long Words & Self-implemented \\ \cline{2-5} 
 & \multirow{10}{*}{\rotatebox{90}{Sensationalism (13)}} & \multirow{4}{*}{Sentiments (7)} & \#/\% Positive Words & \multirow{3}{*}{LIWC} \\ \cline{4-4} 
 &  &  & \#/\% Negative Words &  \\ \cline{4-4} 
 &  &  & Overall \#/\% Emotional Words &  \\ \cline{4-5} 
 &  &  & Avg. Sentiment Score of Words & \begin{tabular}[c]{@{}c@{}}NLTK.Sentiment\\ Package\end{tabular} \\ \cline{3-5} 
 &  & \multirow{4}{*}{Punctuations (4)} & \# `!' & Self-implemented \\ \cline{4-5} 
 &  &  & \# `?' & Self-implemented \\ \cline{4-5} 
 &  &  & \# `...' & Self-implemented \\ \cline{4-5} 
 &  &  & Overall \# `!' `?' `...' & Self-implemented \\ \cline{3-5} 
 &  & \multirow{2}{*}{\begin{tabular}[c]{@{}c@{}}Similarity between\\ Headline \& Bodytext (2)\end{tabular}} & Word2Vec + Cosine Distance & \cite{mikolov2013efficient} \\ \cline{4-5} 
 &  &  & Sentence2Vec + Cosine Distance & \cite{arora2016simple} \\ \cline{2-5} 
 & \multirow{11}{*}{\rotatebox{90}{News-worthiness (20)}} & \multirow{5}{*}{Quality (8)} & Word2Vec + Cosine Distance & \cite{mikolov2013efficient} \\ \cline{4-5} 
 &  &  & Sentence2Vec + Cosine Distance & \cite{arora2016simple} \\ \cline{4-5} 
 &  &  & \#/\% Content Words & \multirow{2}{*}{LIWC} \\ \cline{4-4}
 &  &  & \#/\% Function Words &  \\ \cline{4-5} 
 &  &  & \#/\% Stop Words & Self-implemented \\ \cline{3-5} 
 &  & \multirow{6}{*}{Informality (12)} & \#/\% Swear Words & \multirow{6}{*}{LIWC} \\ \cline{4-4}
 &  &  & \#/\% Netspeak &  \\ \cline{4-4}
 &  &  & \#/\% Assent &  \\ \cline{4-4}
 &  &  & \#/\% Nonfluencies &  \\ \cline{4-4}
 &  &  & \#/\% Fillers &  \\ \cline{4-4}
\multirow{-34}{*}{\rotatebox{90}{\textbf{Clickbait-related Attributes (CBAs) (44)}}} &  &  & Overall \#/\% Informal Words &  \\ \hline
\end{longtable}
\end{small}
\end{appendix}

\end{document}